\documentclass{article}
\usepackage{spconf,amsmath,graphicx}

\usepackage{multirow}
\usepackage{booktabs} 

\usepackage{xspace}
\usepackage{color}
\usepackage{multirow}
\usepackage{graphics}
\usepackage{adjustbox}
\usepackage{subfigure}
\usepackage{amsmath}
\usepackage{paralist} 
\usepackage{enumitem} 
\usepackage{booktabs} 
\usepackage{array}
\usepackage{grffile}
\usepackage{diagbox}
\usepackage{mathtools}

\usepackage[figuresright]{rotating}
\usepackage{makecell}
\usepackage{tabulary}

\graphicspath{{figures/}}

\usepackage[pagebackref=false,breaklinks=true,letterpaper=true,colorlinks,bookmarks=false,citecolor=blue,linkcolor=blue]{hyperref}

\newcommand{\Paragraph}[1]{{\vspace{-2mm}\flushleft\textbf{#1}}} 

\long\def\ignorethis#1{}

\usepackage{tabularx}

\definecolor{B1}{RGB}{237,219,201}
\definecolor{MyBlueb}{RGB}{95,169,236}
\definecolor{MyOrange}{RGB}{255,177,98}
\definecolor{MyReda}{RGB}{193,39,45}
\definecolor{gray}{rgb}{0.5,0.5,0.5}
\definecolor{MyBlue}{rgb}{0,0,1.0}
\definecolor{MyYellow}{rgb}{0.9,0.9,0}
\definecolor{MyRed}{rgb}{0.8,0.2,0}
\definecolor{MyGreen}{rgb}{0,0.5,0.0}
\definecolor{MyGray}{rgb}{0.4,0.4,0.4}

\newlength\paramargin
\newlength\figmargin
\newlength\secmargin

\newcolumntype{L}[1]{>{\raggedright\let\newline\\\arraybackslash\hspace{0pt}}m{#1}}
\newcolumntype{C}[1]{>{\centering\let\newline\\\arraybackslash\hspace{0pt}}m{#1}}
\newcolumntype{R}[1]{>{\raggedleft\let\newline\\\arraybackslash\hspace{0pt}}m{#1}}

\def\eg{e.g.,~}

\newcommand{\secref}[1]{Section~\ref{#1}}
\newcommand{\figref}[1]{Figure~\ref{#1}}
\newcommand{\tabref}[1]{Table~\ref{#1}}
\newcommand{\eqnref}[1]{Equation~\eqref{#1}}

\makeatletter

\setbox0\hbox{$\xdef\scriptratio{\strip@pt\dimexpr
		\numexpr(\sf@size*65536)/\f@size sp}$}

\newcommand{\scriptveryshortarrow}[1][5pt]{{%
		\hbox{\rule[\scriptratio\dimexpr\fontdimen22\textfont2-.2pt\relax]
			{\scriptratio\dimexpr#1\relax}{\scriptratio\dimexpr.4pt\relax}}%
		\mkern-5mu\hbox{\let\f@size\sf@size\usefont{U}{lasy}{m}{n}\symbol{41}}}}

\makeatother

\title{Enhanced Deep Animation Video Interpolation}
\name{Wang Shen$^\star$, Cheng Ming$^\star$, Wenbo Bao$^\dagger$, Guangtao Zhai$^\star$, Li Chen$^\star$, Zhiyong Gao$^\star$\thanks{Corresponding author: Guangtao Zhai. Codes and supplementary material are at: \url{https://github.com/laomao0/AutoSktFI} }}
\address{$^\star$ Shanghai Jiao Tong University \quad\quad$^\dagger$ PointSpread Technology Inc.}
\begin{document}
%
\maketitle
\begin{abstract}
    Existing learning-based frame interpolation algorithms extract consecutive frames from high-speed natural videos to train the model.
    Compared to natural videos, cartoon videos are usually in a low frame rate.
    Besides, the motion between consecutive cartoon frames is typically nonlinear, which breaks the linear motion assumption of interpolation algorithms.
    Thus, it is unsuitable for generating a training set directly from cartoon videos.
    For better adapting frame interpolation algorithms from nature video to animation video, we present AutoFI, a simple and effective method to automatically render training data for deep animation video interpolation.
    AutoFI takes a layered architecture to render synthetic data, which ensures the assumption of linear motion.
    Experimental results show that AutoFI performs favorably in training both DAIN and ANIN.
    However, most frame interpolation algorithms will still fail in error-prone areas, such as fast motion or large occlusion.
    Besides AutoFI, we also propose a plug-and-play sketch-based post-processing module, named SktFI, to refine the final results using user-provided sketches manually.
    With AutoFI and SktFI, the interpolated animation frames show high perceptual quality.
\end{abstract}

\begin{figure}[t]
	\footnotesize
	\centering	
	\includegraphics[width=0.95\linewidth]{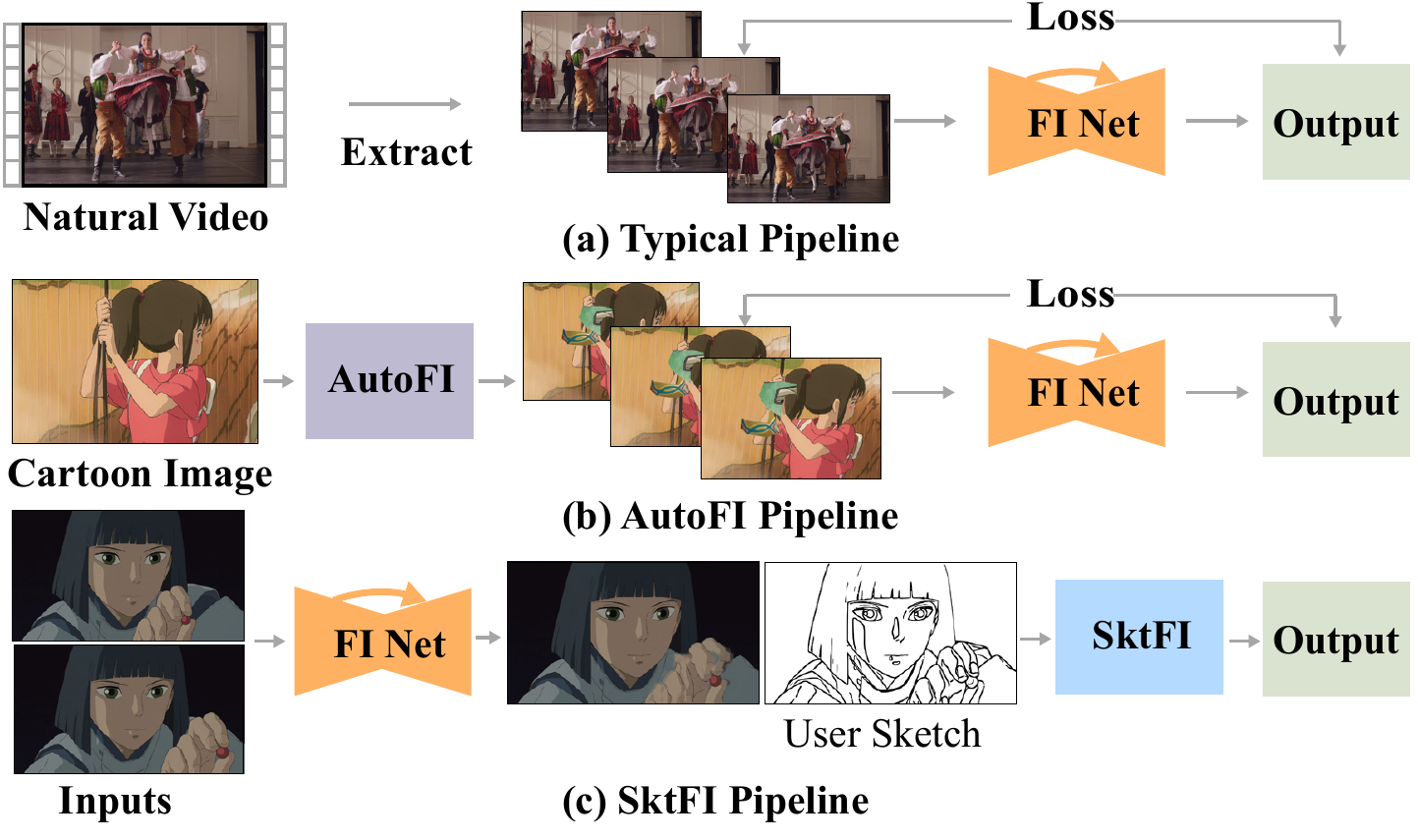}
	\vspace{-10pt}
	\caption{
        Pipelines for video frame interpolation.
        A typical pipeline in (a) extracts three consecutive frames from a natural video.
        The first and the third frames are inputs for the interpolation network, and the middle one is used as the ground-truth frame.
        AutoFI in (b) uses a single background cartoon image and masked layers to synthesize triplet frames with linear motions.
        SktFI in (c) utilizes user-provided sketch to rectify error-prone areas in frame interpolation algorithms.
	    } 
	\label{fig:pipeline} 
	\vspace{-15pt}
\end{figure}

\section{Introduction}\label{sec:intro}
    Learning-based frame interpolation algorithms boot numerous applications, such as slow motion generation~\cite{jiang2018super}, neural view rendering~\cite{niklaus2020softmax}, frame rate up-conversion~\cite{bao2019depth,shen2020blurry,shen2020video,shen2021spatial}.
    Datasets are the critical driving force for training a frame interpolation algorithm.
    Most state-of-the-art frame interpolation algorithms use extracted consecutive frames from high-speed natural videos for training, such as dataset Vimeo90K~\cite{xue2019video}, Adobe240~\cite{su2017deep}.
    Compared to natural videos, cartoon videos usually have low frame rates. 
    In the animation industry, drawing an in-between frame cost complex and enormous labor. 
    Producers usually replicate one frame two or three times to save cost.
    Besides, unlike nature videos, the motion between consecutive cartoon frames is typically nonlinear, which breaks the linear motion assumption for frame interpolation~\cite{bao2019depth,jiang2018super}.
    Thus, it is unsuitable to directly generate a training set from cartoon videos to adapt frame interpolation algorithms from nature video to animation video.
    
    ANIN~\cite{siyao2021deep} is the most related work that considers animation video frame interpolation using a deep neural network.
    ANIN proposes Segment-Guided Matching to address the cartoon frame alignment problem.
    Their training set is manually selected from cartoon frames that appear to move linearly.
    However, manual selection severely limits the scalability of training and can not avoid the non-linearity, which misleads the frame synthesis network, resulting in blurry interpolated frames, as shown in~\figref{fig:compare_autofi_1}.

    In this paper, we first present AutoFI (see~\figref{fig:pipeline}(b)), a simple and effective method to \textbf{Auto}matically render training data for deep animation video \textbf{F}rame \textbf{I}nterpolation.
    AutoFI takes a layered architecture to render three consecutive cartoon frames using a single background image and masked layers.
    AutoFI ensures the basic assumption of linear motion, the interpolation network trained using AutoFI (\eg DAIN~\cite{bao2019depth} and ANIN~\cite{siyao2021deep}) presents clear and high-quality content.
    However, most frame interpolation algorithms will still fail at error-prone areas, such as fast motion or large occlusion.
    Besides AutoFI, we also propose a plug-and-play \textbf{Sk}e\textbf{t}ch-based \textbf{F}rame \textbf{I}nterpolation module, named SktFI (see~\figref{fig:pipeline}(c)), to manually refine the final results using user-provided sketches.
    With AutoFI and SktFI, the interpolated animation frames show high perceptual quality.
    In this paper, we have the following contributions: 1) we first present AutoFI, an effective method for arbitrary rendering training data for animation in-between; 2) we introduce SktFI, a practical post-processing module to rectify error-prone areas in frame interpolation; 3) Numerical and visual results demonstrate the effectiveness of our method.
    
\section{Methodology}

\subsection{AutoFI Pipeline}\label{sec:autofi}
    Given two animation frames $\mathbf{I}_1$ and $\mathbf{I}_3$, the goal of frame interpolation is to synthesize an intermediate frame $\hat{\mathbf{I}}_{2}$.
    The training set for learning-based frame interpolation algorithms typically contains three frames $\mathbf{I}_1$, $\mathbf{I}_2$  and $\mathbf{I}_3$.
    $\mathbf{I}_1$ and $\mathbf{I}_3$ are used as input and  $\mathbf{I}_2$ is used as the ground-truth frame for supervised learning.
    AutoFI aims to enhance the perceptual quality of interpolated frames by generating a training set on animation videos to reduce domain bias and then training frame interpolation.
    AutoFI takes a layered approach~\cite{wang1994representing,sun2010layered,sun2021autoflow} to synthesize consecutive frames $\mathbf{I}_1$, $\mathbf{I}_2$ and $\mathbf{I}_3$ by using one background frame and several masked layered contents.

\begin{figure}[t]
	\footnotesize
	\centering	
	\includegraphics[width=0.9\linewidth]{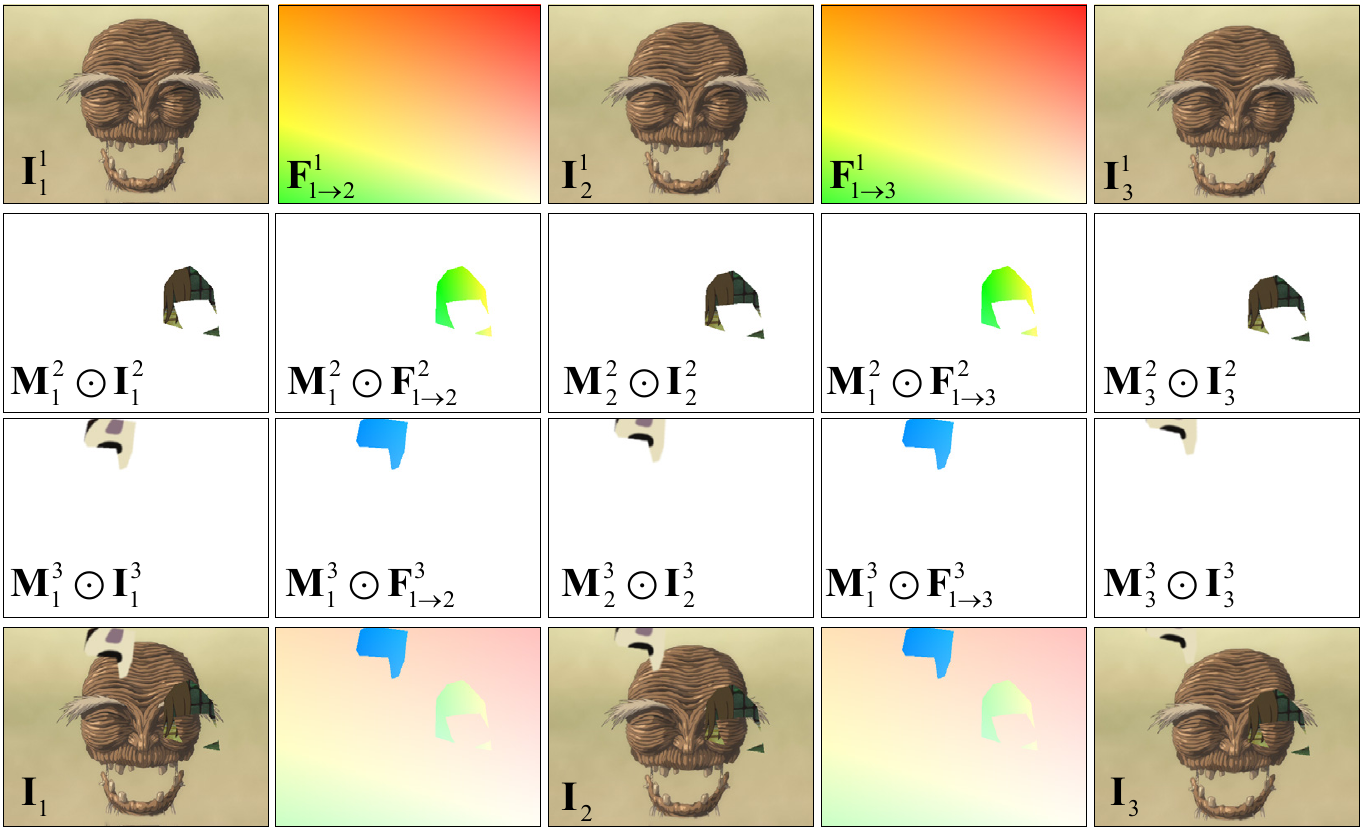}
	\vspace{-10pt}
	\caption{
        AutoFI pipeline.
        Using $K=2$ as an example.
	    } 
	\label{fig:autofi} 
	\vspace{-8pt}
\end{figure}

\begin{figure}[t]
	\footnotesize
	\centering	
	\includegraphics[width=0.90\linewidth]{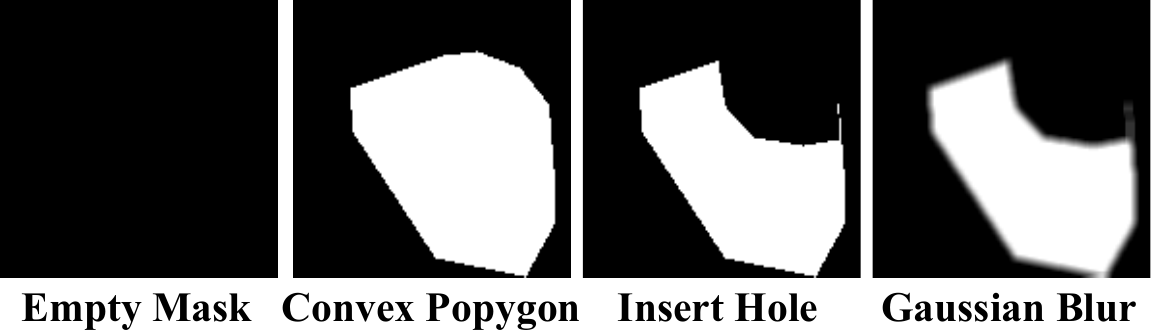}
	\vspace{-10pt}
	\caption{
       Illustration of layer masks with random polygons.
	    } 
	\label{fig:mask} 
	\vspace{-15pt}
\end{figure}

    Here, we introduce the AutoFI pipeline, as shown in~\figref{fig:autofi}.
    We first construct an image dataset $\mathcal{S}$ by independently extracting images from an animation video, such as Spirited Away.
    Notice that the image dataset $\mathcal{S}$ does not need consecutive video frames.
    For the background layer (i.e., the first layer with $k=1$), we randomly select one image $\mathbf{I}_1^{k=1}$ from $\mathcal{S}$, as shown in~\figref{fig:autofi}.
    For upper layers with $k\in[2,K]$, we also randomly select one image from the image dataset $\mathcal{S}$, defined as $\mathbf{I}_{1}^{k}$, where $k\in[2,K]$.
    Besides, we generate binary mask $\mathbf{M}_{1}^{k}$ for upper layers.
    The mask is generated using random polygons, as shown in~\figref{fig:mask}.

    For each layer, we also generate random per-pixel motion field (i.e., optical flow) between consecutive frames using homography transformation, defined by $\mathbf{F}_{1\rightarrow2}^{k}$ and $\mathbf{F}_{1\rightarrow3}^{k}$.
    Notice that under the linear motion assumption, we set:
    \begin{equation}\label{eq:flow}
    \mathbf{F}_{1\rightarrow3}^{k} = 2 \cdot  \mathbf{F}_{1\rightarrow2}^{k}.
    \end{equation}

    The optical flows are used to warp $\mathbf{I}_1^{k}$ to the time index $n$:
    \begin{align}
         \mathbf{I}_{n}^k &= \mathcal{W}(\mathbf{I}_1^{k}, \mathbf{F}_{1\rightarrow n}^k),  \quad \quad  k \in [1,K], n \in \{2,3\}, \\
        \mathbf{M}_{n}^k &= \mathcal{W}(\mathbf{M}_1^{k}, \mathbf{F}_{1\rightarrow n}^k),  \quad  k \in [1,K], n \in \{2,3\},
    \end{align}
    where $\mathcal{W}(\cdot)$ is the forward-warping function, which we adopt SoftmaxSplat\cite{niklaus2020softmax} with the average splatting mode.

    We then conduct an iterative layered approach to construct the final frame by iterating from $k=1$ to $k=K$:
    \begin{align}
        \mathbf{I}_n &:= (\mathbf{1}-\mathbf{M}_n^k) \cdot  \mathbf{I}_n + \mathbf{M}_n^k \cdot \mathbf{I}_n^k,k\in [1,K], n \in[1,3],
    \end{align}
    where the initial values for the background layer (i.e., $k=1$) are set $\mathbf{I}_1=\mathbf{0}$ and $\mathbf{M}_{n}^1=\mathbf{1}$.
    Notice that we could also obtain ground-truth optical flows by:
     \begin{align}
        \mathbf{F}_{1\rightarrow n} &:= (\mathbf{1}-\mathbf{M}_n^k) \cdot  \mathbf{F}_{1 \rightarrow n} + \mathbf{M}_n^k \cdot \mathbf{F}_{1\rightarrow n}^k,
    \end{align}
    where $k\in [1,K], n \in[2,3]$.
    However, this paper does not utilize the flows for training frame interpolation models.
    The ground-truth flows could facilitate related optical flow estimation algorithms for future research~\cite{sun2018pwc,teed2020raft}.

    \Paragraph{Layer Mask.}\label{sec:mask}
    We generate random convex polygons~\cite{sun2021autoflow} as binary masks for the upper layers with $k\in[2, K]$ to simulation object movements.
    The synthesized polygons have a random number of sides, and we randomly make the hole by nesting a small polygon into it, as shown in~\figref{fig:mask}.
    Similar to the previous work\cite{sun2021autoflow}, we apply a Gaussian Blur filter on both the layer masks and masked objects.
    The standard deviation $\delta$ of Gaussian blur is calculated by evaluating the average flow magnitude over the mask:
    \begin{equation}
        \delta = \text{log} \big( \frac{\text{sum}(||\mathbf{M}_n^k \cdot \mathbf{F}_{1\rightarrow n}^k||)}{\text{sum}(|| \mathbf{M}_n^k||) \cdot \alpha }  \big),
    \end{equation}
    where scaling factor $\alpha$ is set as $0.1$ empirically.

    \Paragraph{Motion Field.}\label{sec:motion}
    We use random homography transformation to synthesize optical flows (see~\eqnref{eq:flow}), which combines several types of motion, including scale, rotation, translation, and perspective motion.
    We first generate a random homography matrix, denoted by $\mathbf{H}$.
    We then compute the transformed pixel location by:
    \begin{equation}
        \mathbf{y}(p) = \mathbf{H} \cdot \mathbf{x}(p),
    \end{equation}
    where $\mathbf{x}(p)$ is the pixel location of pixel $p$ and $\mathbf{y}(p)$ refers to the transformed location.
    We compute the motion vector by:
    \begin{equation}
        \mathbf{f}_{x \rightarrow y}(p) =   \mathbf{y}(p) - \mathbf{x}(p).
    \end{equation}
    The optical flow field $\mathbf{F}_{1\rightarrow n}$ is a set collecting all pixels $p$, defined by $\mathbf{F}_{1\rightarrow n} = \{ \mathbf{f}_{1 \rightarrow n}(p)~|~\forall p \}$.
    Notice that here we drop superscript $k$ for  $\mathbf{F}_{1\rightarrow n}^k$ for brevity.

\subsection{SktFI Pipeline}\label{sec:skefi}
    
    \begin{figure}[t]
	\footnotesize
	\centering	
	\includegraphics[width=0.9\linewidth]{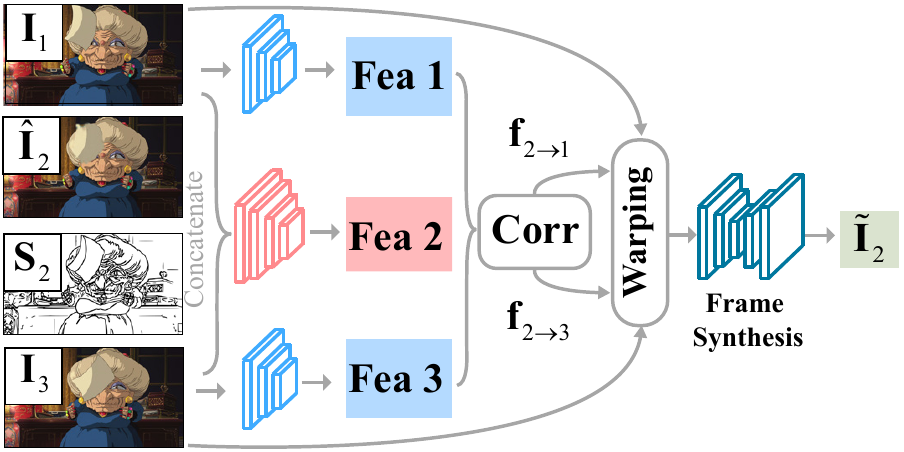}
	\vspace{-10pt}
	\caption{
	   Framework of the proposed SktFI network.
	    } 
	\label{fig:sktfi} 
	\vspace{-10pt}
\end{figure}

    SktFI aims to refine the initial interpolated image $\mathbf{\hat{I}}_2$ using a user-provided sketch $\mathbf{S}_2$ as a reference, as shown in~\figref{fig:sktfi}.
    The sketch $\mathbf{S}_2$ could be generated from the initial interpolated frame using sketch simplification algorithms~\cite{simo2018mastering}, then the user refines the error-prone synthesized areas.
    
    We first extract features from input $\mathbf{I}_1$ and $\mathbf{I}_3$ using feature extractor, denoted by $\{\mathbf{Fea}_1, \mathbf{Fea}_3\}$.
    We the use another extend extractor to extract feature $\mathbf{Fea}_2$ from concatenated frames $\{\mathbf{I}_1, \mathbf{\hat{I}}_2, \mathbf{S}_2, \mathbf{I}_3\}$.
    Then the extracted frames are forward into correlation module from~\cite{sun2018pwc} to synthesize bidirectional optical flows, denoted by $\mathbf{f}_{2\rightarrow 1}$ and  $\mathbf{f}_{2\rightarrow 3}$.
    Finally, we backward warping $\{\mathbf{I}_1, \mathbf{\hat{I}}_2, \mathbf{S}_2, \mathbf{I}_3\}$ guided by estimated flows $\mathbf{f}_{2\rightarrow 1}$ and  $\mathbf{f}_{2\rightarrow 3}$ into the intermediate time index, and the a frame synthesis module takes all the warped frames as input to synthesize the refined image $\mathbf{\tilde{I}}_2$.
    The supplementary material provides network details of feature extractors and the frame synthesis module.

\section{Experiments}

\subsection{Dataset}
    The training data for AutoFI is generated using an animation film (see~\secref{sec:autofi}), called Spirited Away.
    We synthesize 40K triples with resolution $512 \times 384$.
    The training data for SktFI requires user-provided sketches as inputs.
    We use the triplet frames from the training set of ATD-12K~\cite{siyao2021deep} and use sketch generation pipeline from~\cite{portenier2018faceshop} to synthesize the middle frame sketch. 
    The test set is sampled from ATD-12K.
    ATD-12K test set includes several files, and we select the frames that contain Spirited Away as our testing data, which contains 313 triples with resolution $960 \times 540$.
    We use PSNR and SSIM as well as LPIPS~\cite{zhang2018unreasonable} for quantitative comparisons.

\subsection{Loss Function}
    We aim to enhance frame interpolation networks DAIN~\cite{bao2019depth} (trained on natural video set Vimeo90K~\cite{xue2019video}) and ANIN~\cite{siyao2021deep} (trained on manually collected animation dataset ATD-12K~\cite{siyao2021deep}).
    As shown in~\figref{fig:pipeline}(b), we utilize the training set generated based on AutoFI to train DAIN and ANIN using $\mathbf{I}_1, \mathbf{I}_3$ as input.
	We denote the synthesized frame by $\hat{\mathbf{I}}_2$.
	In the AutoFI pipeline, we use pixel reconstruction loss $\mathcal{L}_C$ from~\cite{bao2019depth} to train frame interpolation networks.
	Except for $\mathcal{L}_C$, we also adopt perceptual loss $\mathcal{L}_F$ from~\cite{niklaus2020softmax}  to retain more details in complex cases.
	For example, the model that we fine-tune ANIN~\cite{siyao2021deep} using AutoFI with $\mathcal{L}_F$ is denoted by AutoFI\textit{ -anin -$\mathcal{L}_F$}.
	Notice that AutoFI\textit{ -anin} denotes the model using $\mathcal{L}_C$, unless otherwise specified.
	In the SktFI pipeline, we use pixel reconstruction loss  $\mathcal{L}_C$ to train the refinement network.
	When training the SktFI module, we fix the weights of the pre-stage frame interpolation module.

\begin{figure}[t]
	\footnotesize
	\centering
	\renewcommand{\tabcolsep}{1.0pt} 
	\renewcommand{\arraystretch}{1.0} 
	\begin{tabular}{ccccc}

    \parbox[t]{3.2mm}{\multirow{1}{*}[3.3em]{\rotatebox[origin=c]{90}{{\textit{ -$\mathcal{L}_C$}}}}} &
    \includegraphics[width=0.18\linewidth]{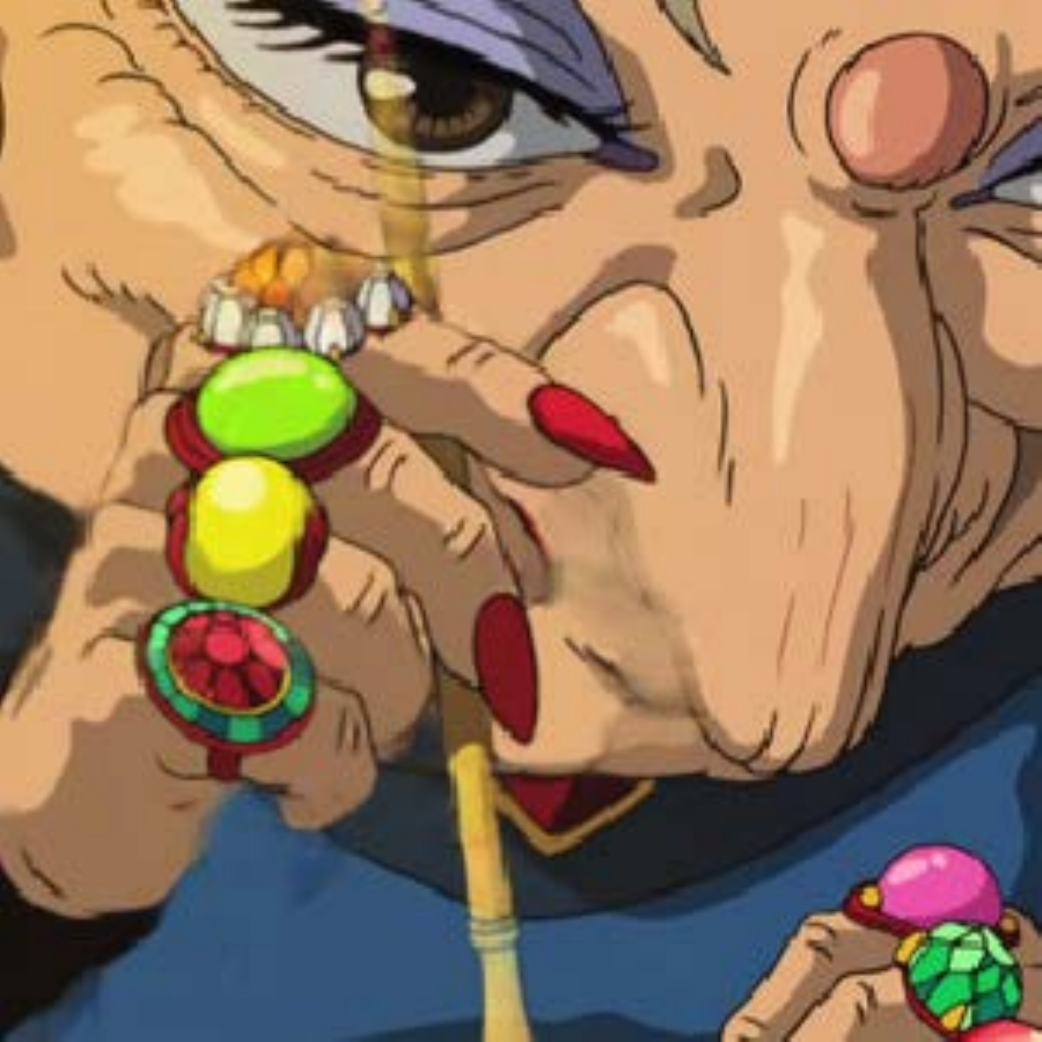} &
    \includegraphics[width=0.18\linewidth]{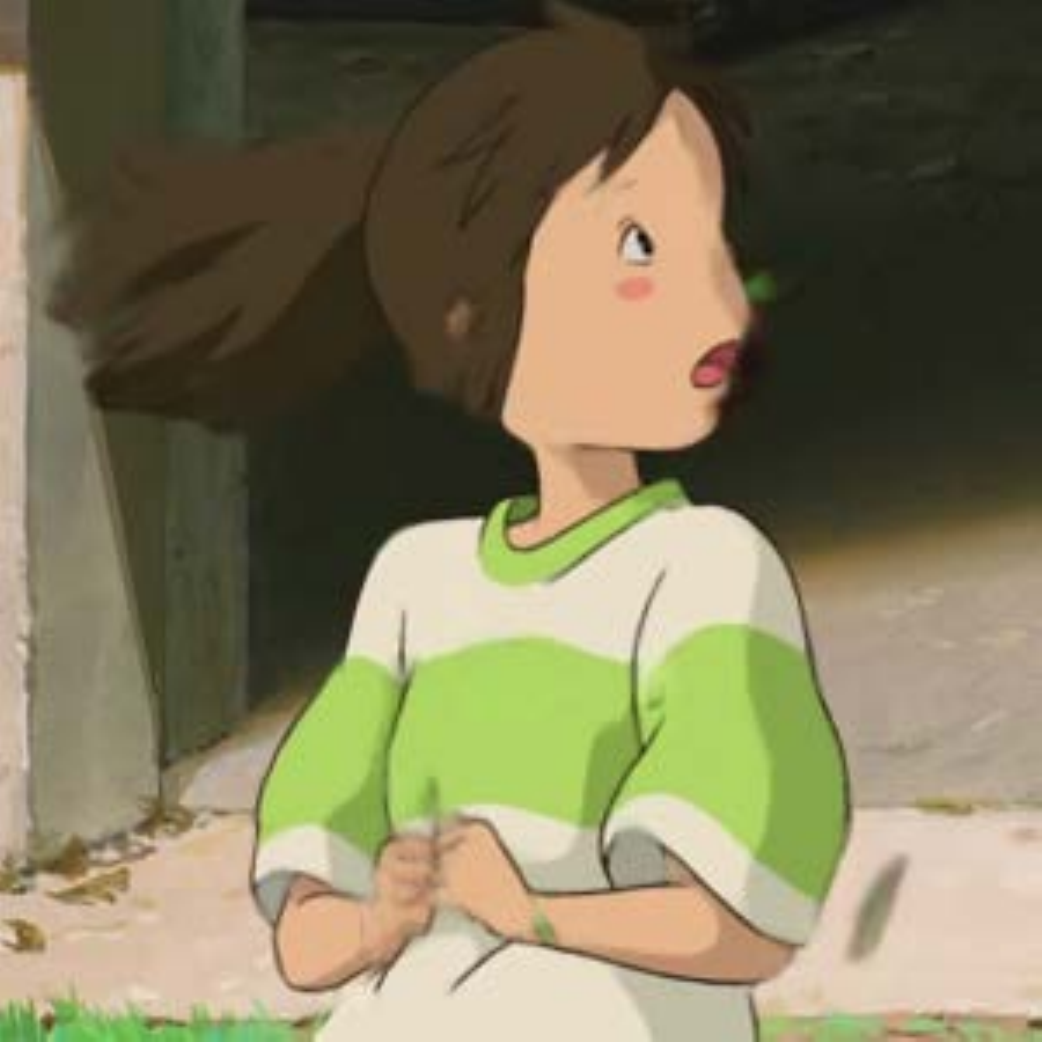} &
    \includegraphics[width=0.18\linewidth]{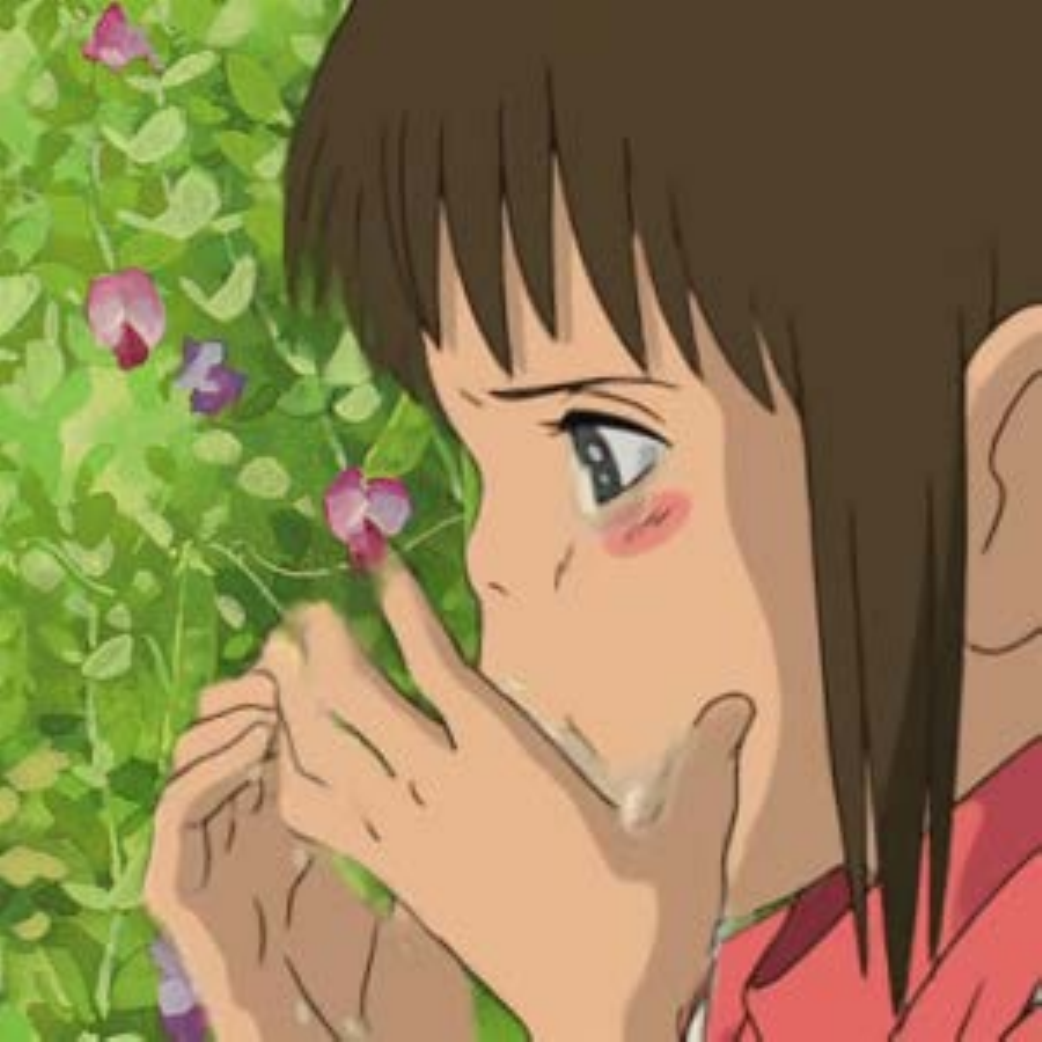} &
    \includegraphics[width=0.18\linewidth]{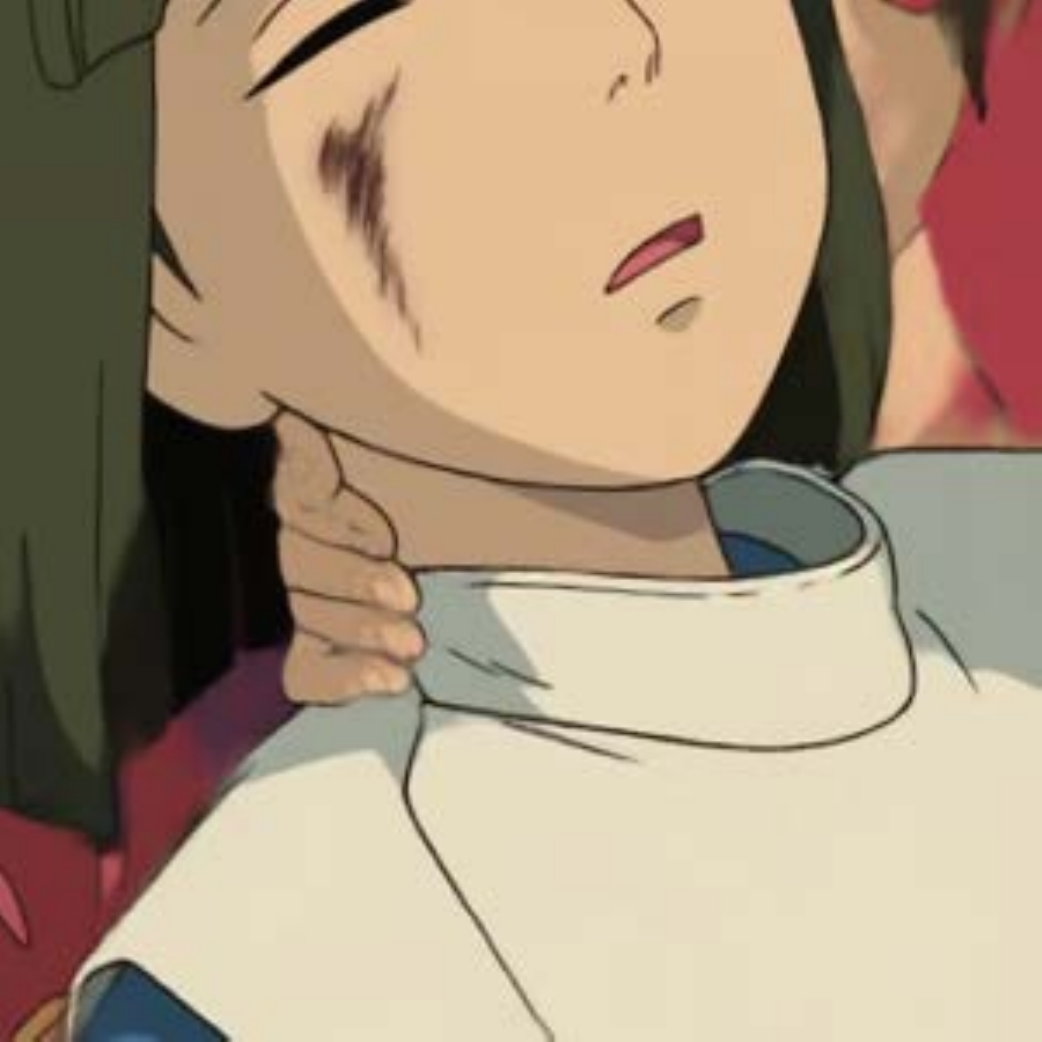} \\
    
    \parbox[t]{3.2mm}{\multirow{1}{*}[3.3em]{\rotatebox[origin=c]{90}{{\textit{ -$\mathcal{L}_F$}}}}} &
    \includegraphics[width=0.18\linewidth]{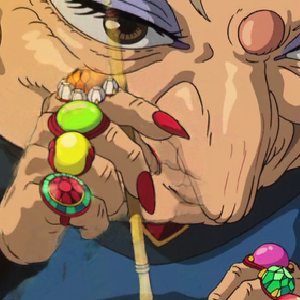} &
    \includegraphics[width=0.18\linewidth]{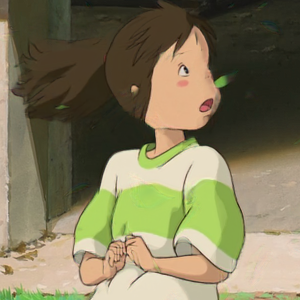} &
    \includegraphics[width=0.18\linewidth]{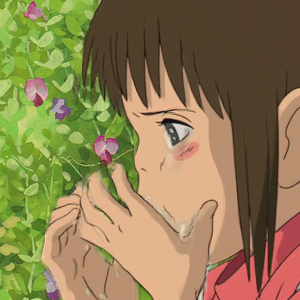} &
      \includegraphics[width=0.18\linewidth]{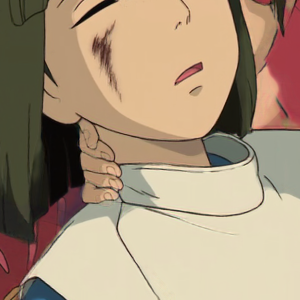} \\
    
\end{tabular}
    \vspace{-10pt}
	\caption{
      Analysis of loss functions on AutoFI\textit{ -anin}.
	}
	\label{fig:abla_percep_loss}
	\vspace{-10pt}
\end{figure}

\begin{figure}[t]
	\footnotesize
	\centering	
	\includegraphics[width=0.80\linewidth]{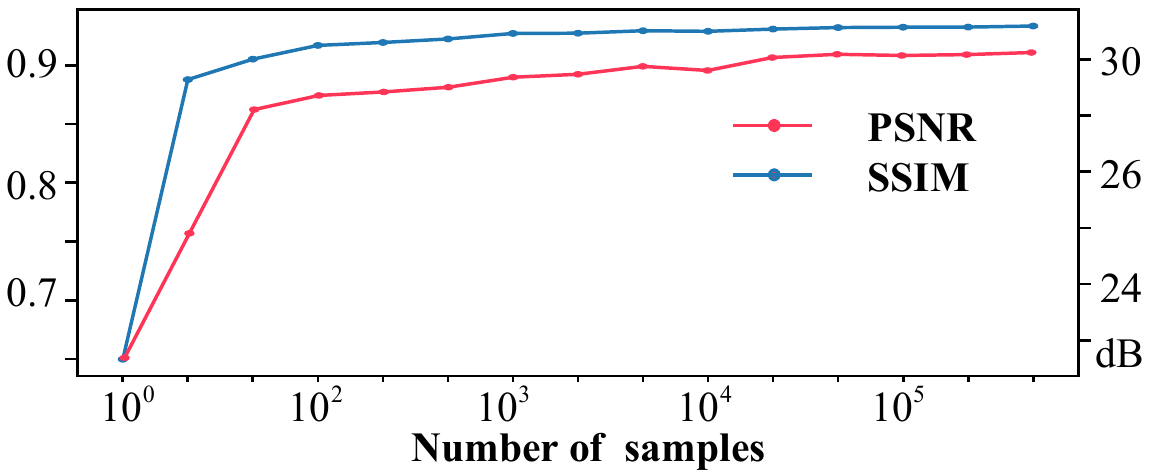}
	\vspace{-10pt}
	\caption{
	   Performance versus training samples of AutoFI dataset (left, SSIM; right, PSNR).
	    } 
	\label{fig:psnr2} 
 	\vspace{-10pt}
\end{figure}

\begin{figure*}[t]
	\footnotesize
	\centering
	\renewcommand{\tabcolsep}{1.0pt} 
	\renewcommand{\arraystretch}{0.5} 
	\begin{tabular}{ccc ccc cc }

    \includegraphics[width=0.105\linewidth]{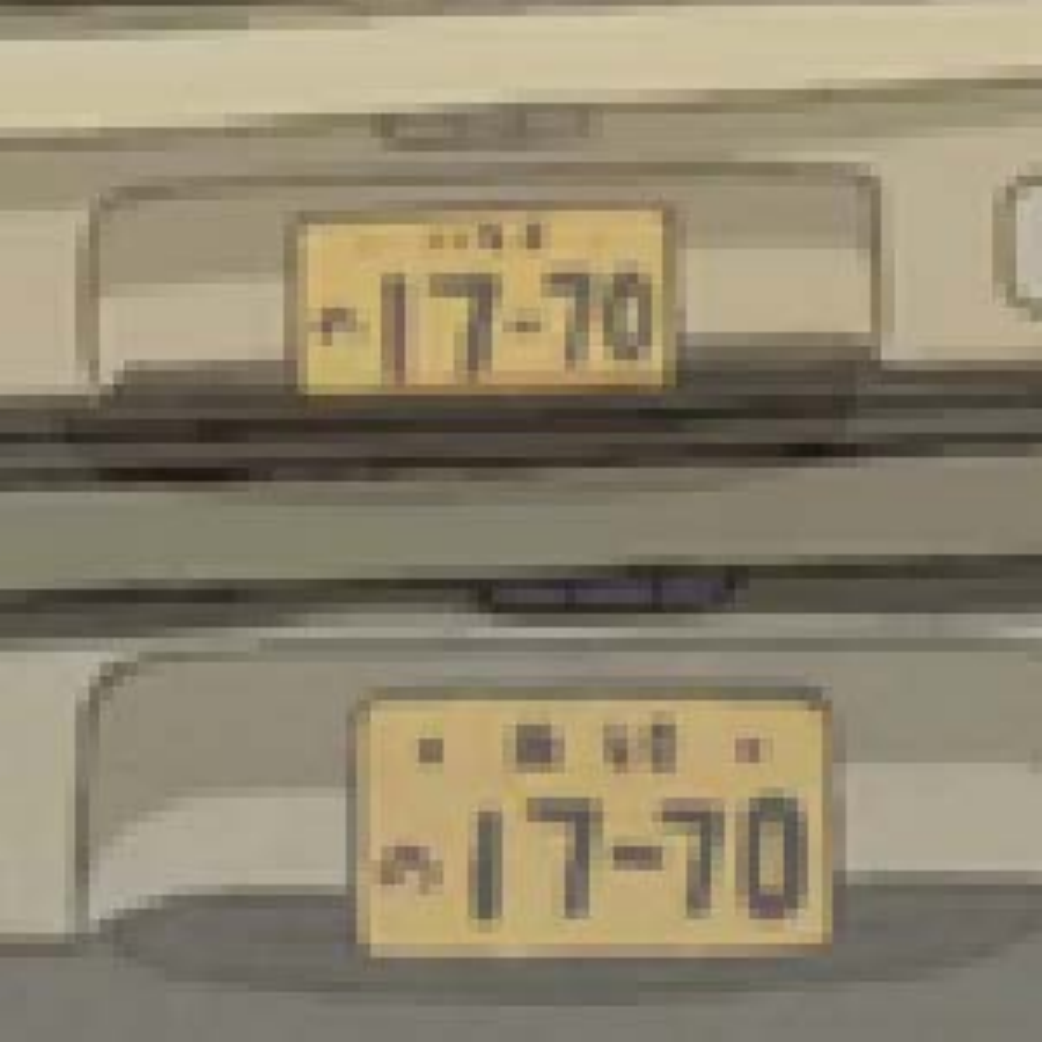}  &
    \includegraphics[width=0.105\linewidth]{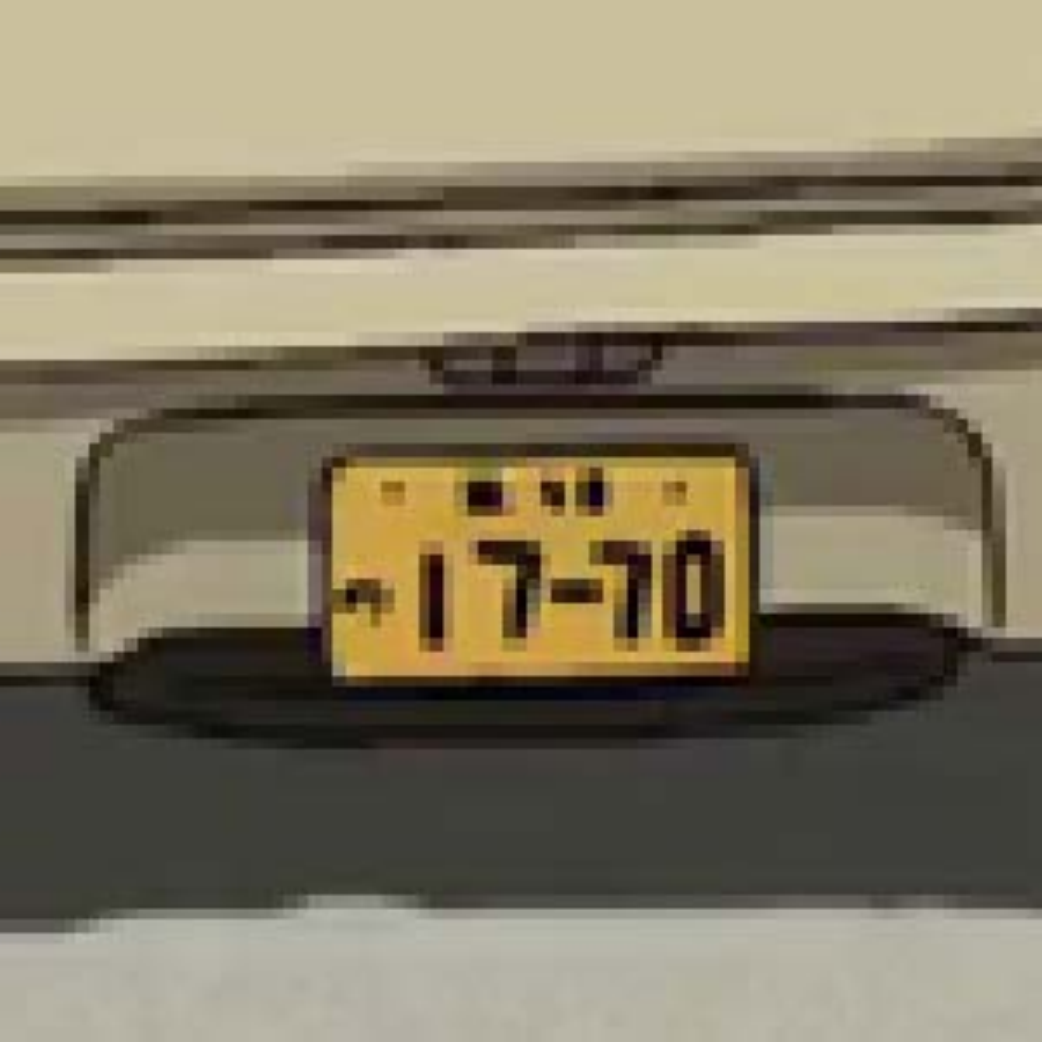} &
    \includegraphics[width=0.105\linewidth]{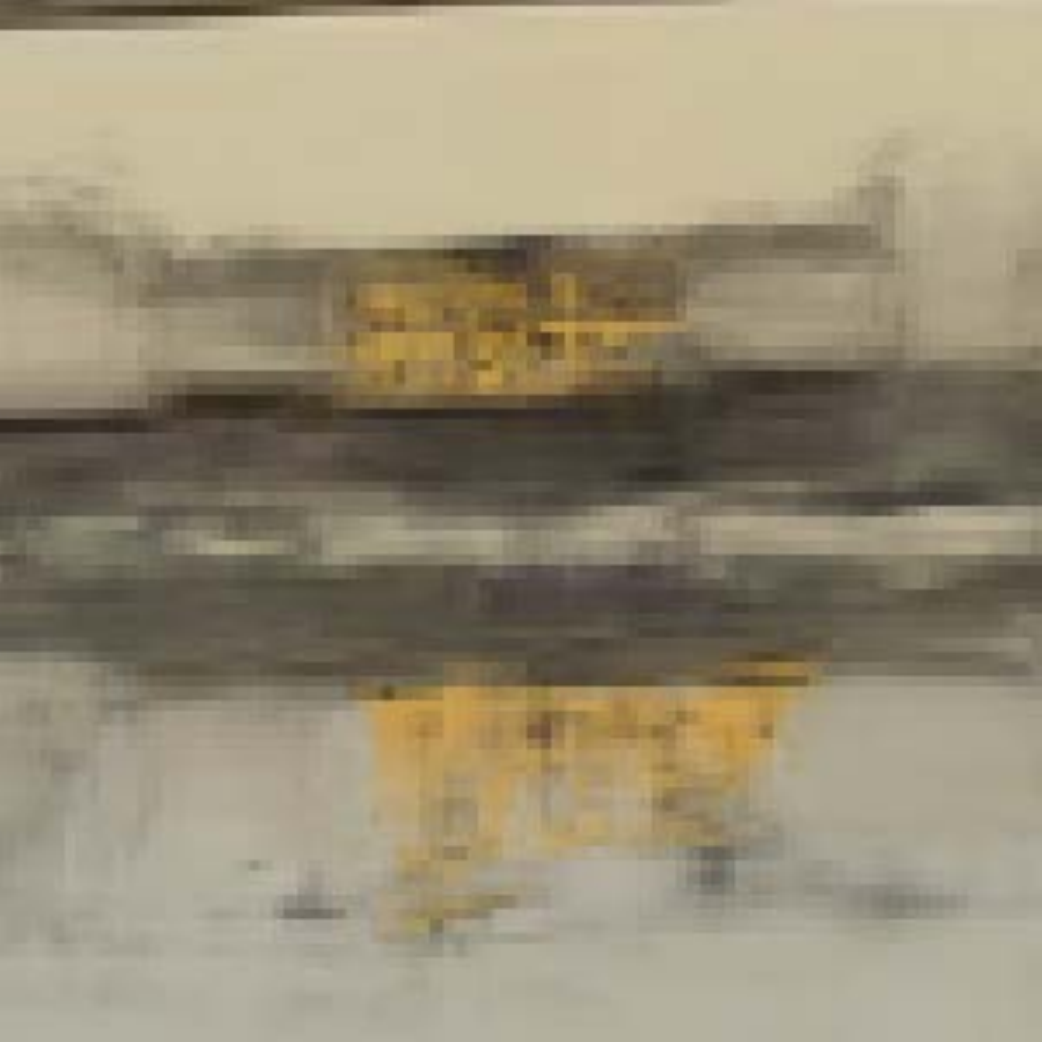}  &
    \includegraphics[width=0.105\linewidth]{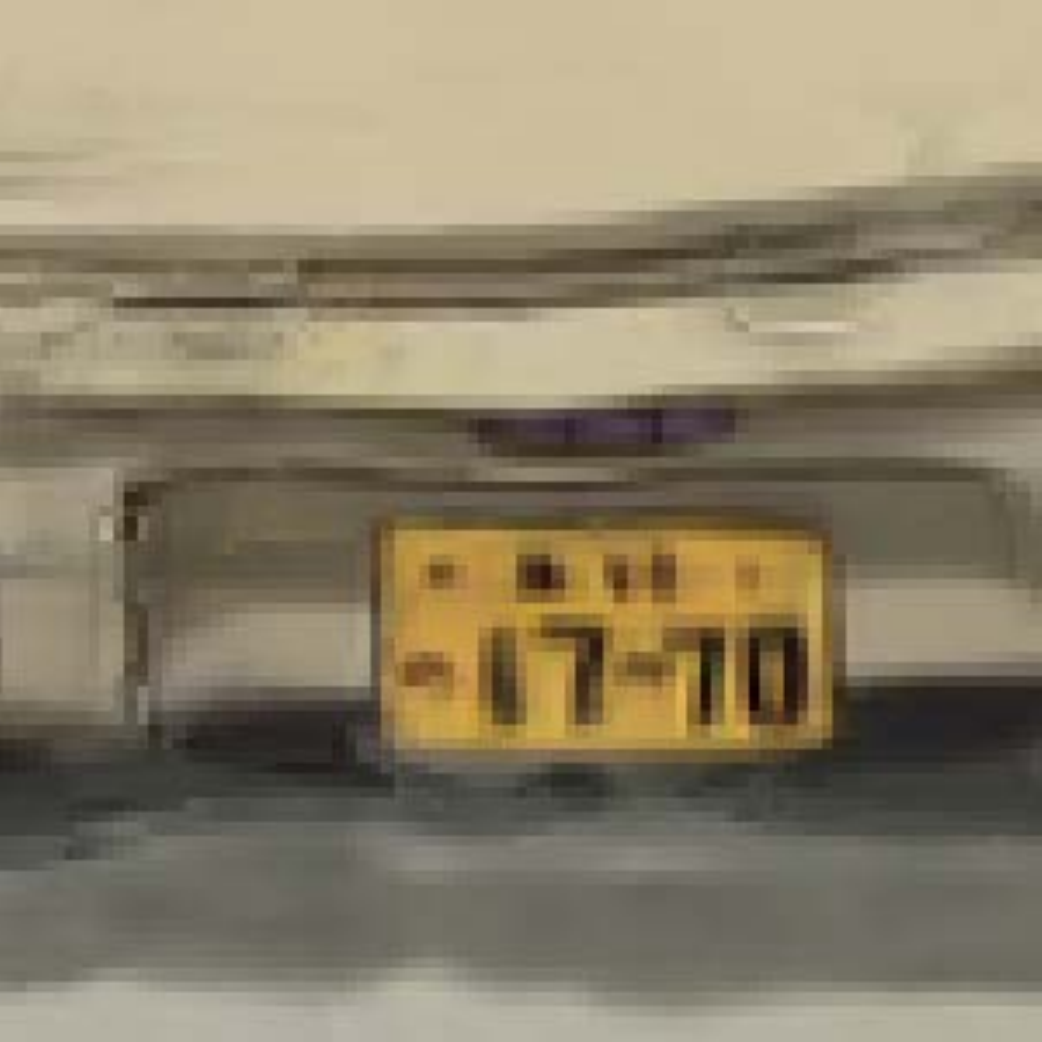}  &
    \includegraphics[width=0.105\linewidth]{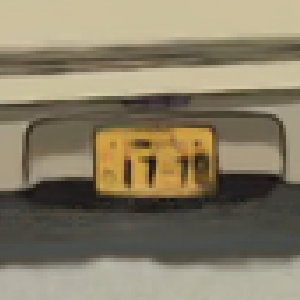}  &
    \includegraphics[width=0.105\linewidth]{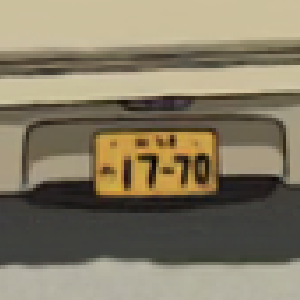}  &
    \includegraphics[width=0.105\linewidth]{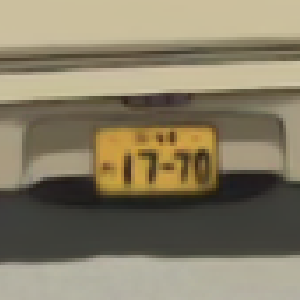}  &
    \includegraphics[width=0.105\linewidth]{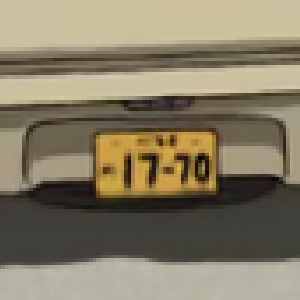} \\
    
    \includegraphics[width=0.105\linewidth]{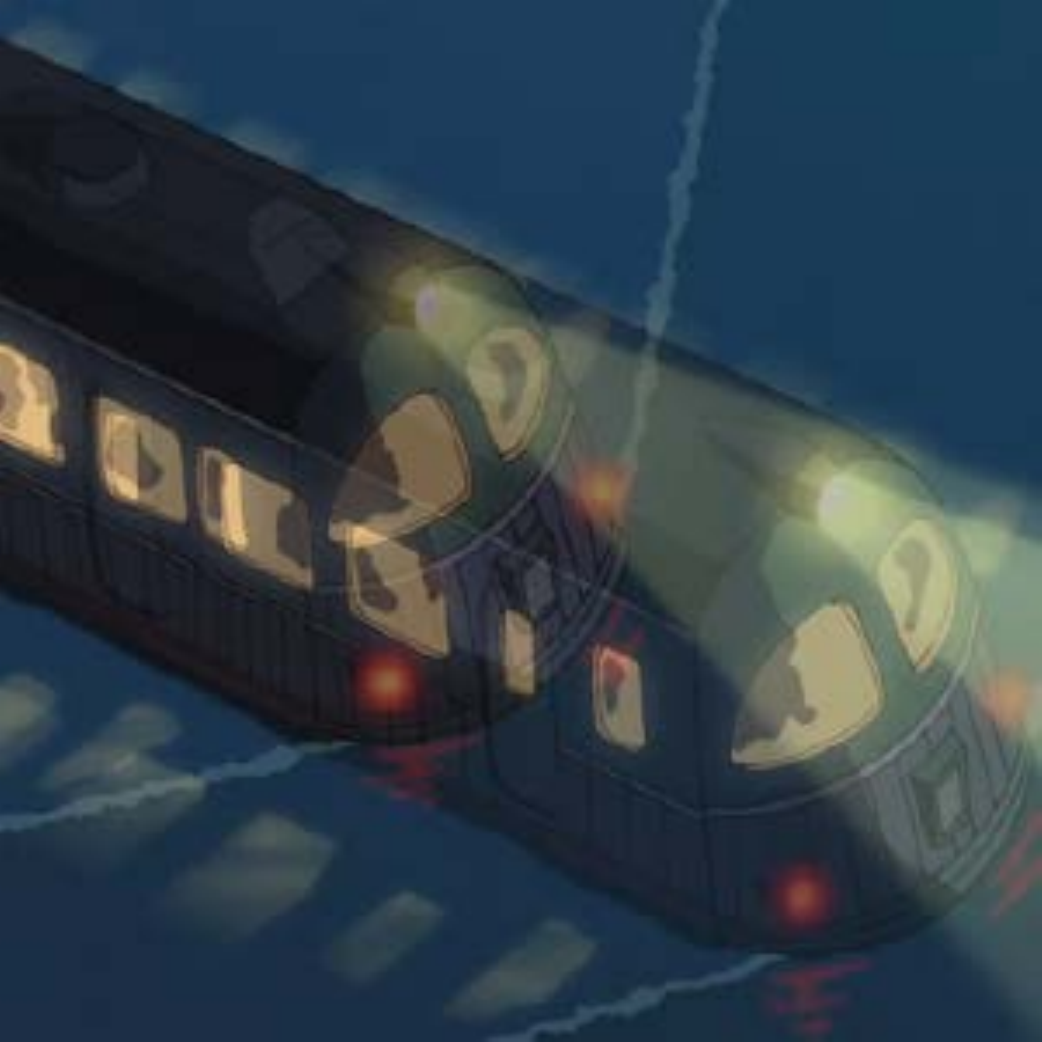}  &
    \includegraphics[width=0.105\linewidth]{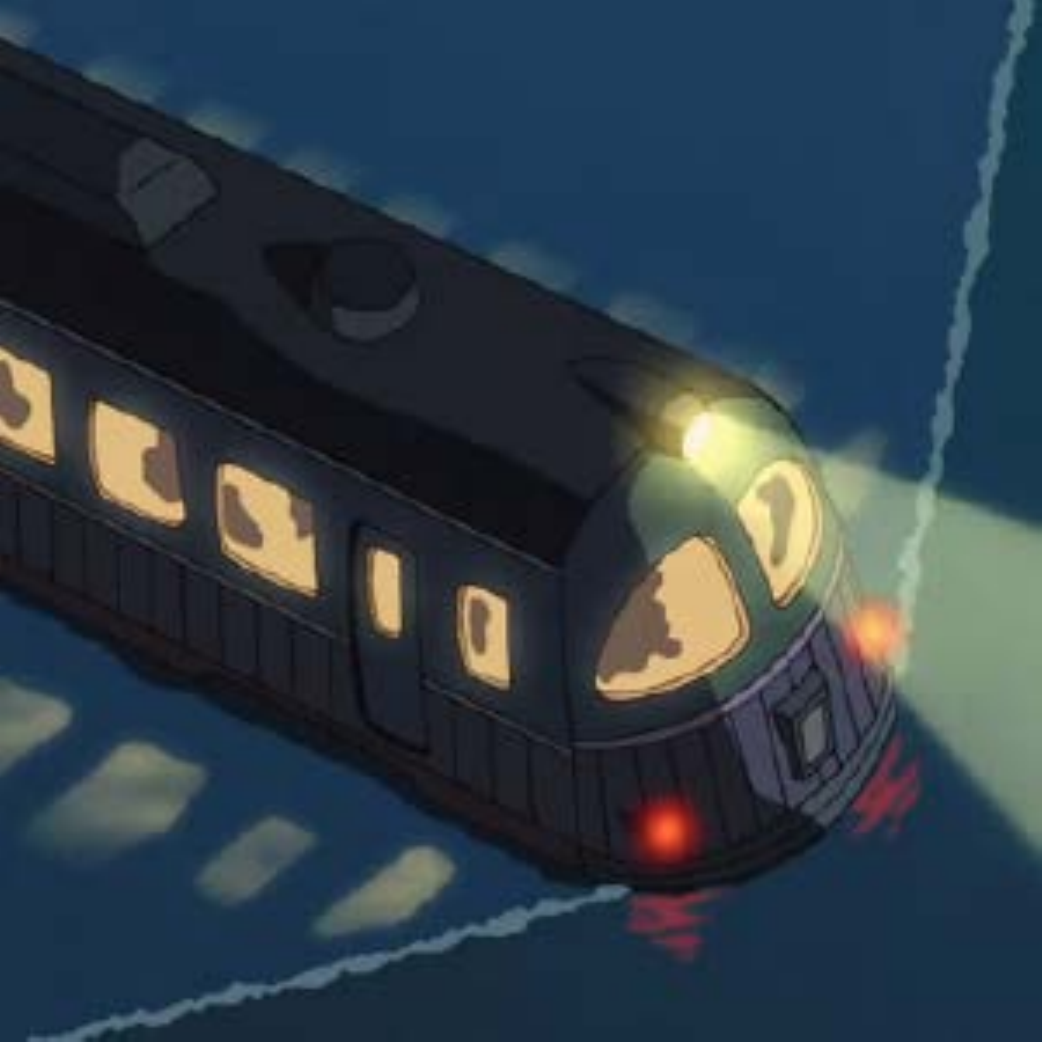} &
    \includegraphics[width=0.105\linewidth]{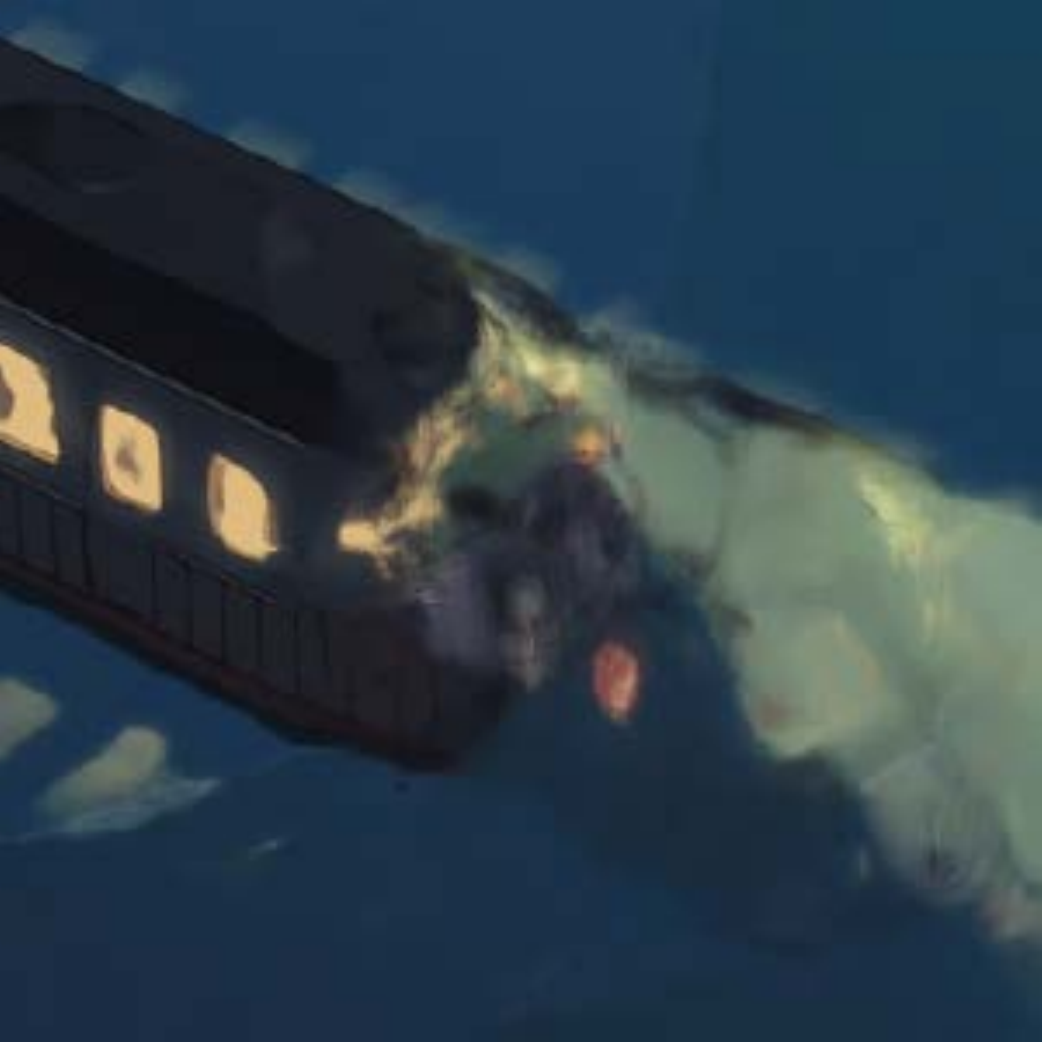}  &
    \includegraphics[width=0.105\linewidth]{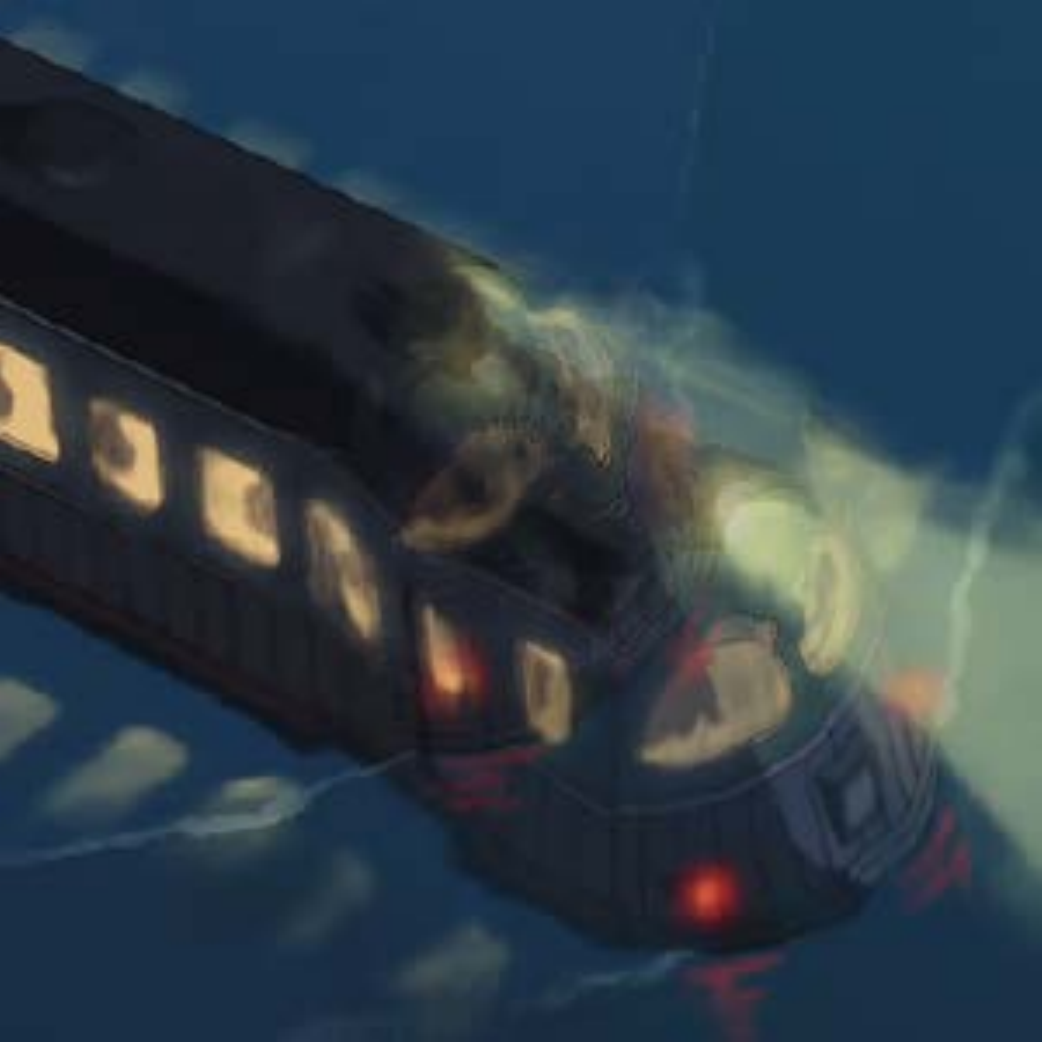}  &
    \includegraphics[width=0.105\linewidth]{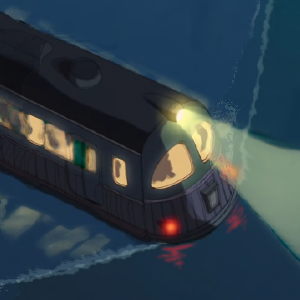}  &
    \includegraphics[width=0.105\linewidth]{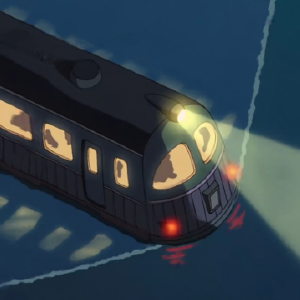}  &
    \includegraphics[width=0.105\linewidth]{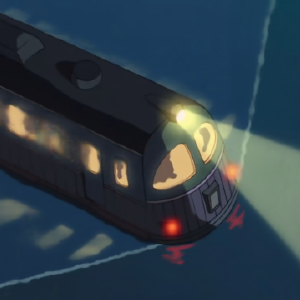}  &
    \includegraphics[width=0.105\linewidth]{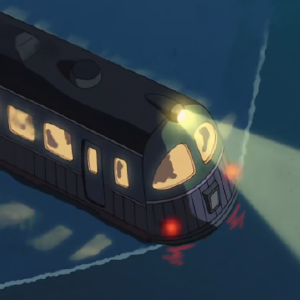}  \\

    \includegraphics[width=0.105\linewidth]{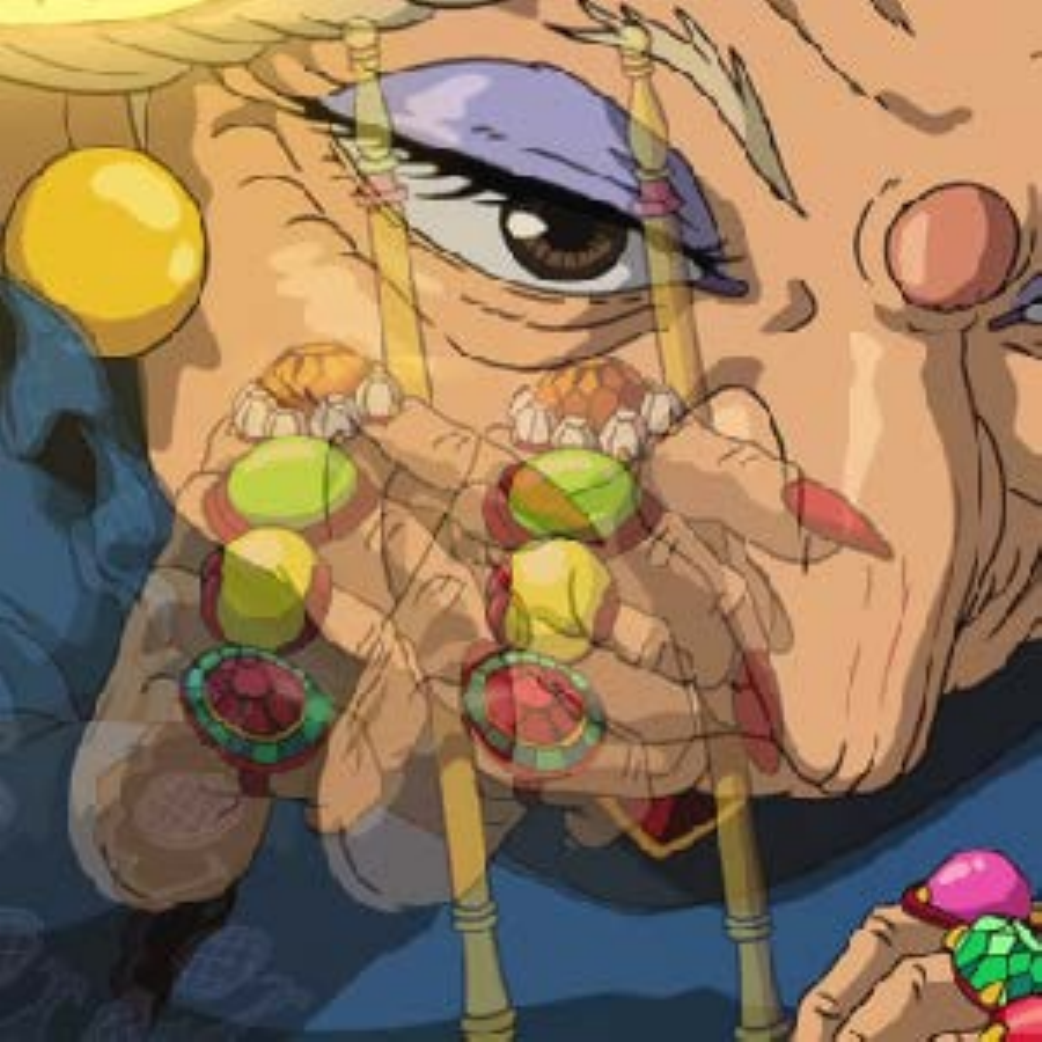}  &
    \includegraphics[width=0.105\linewidth]{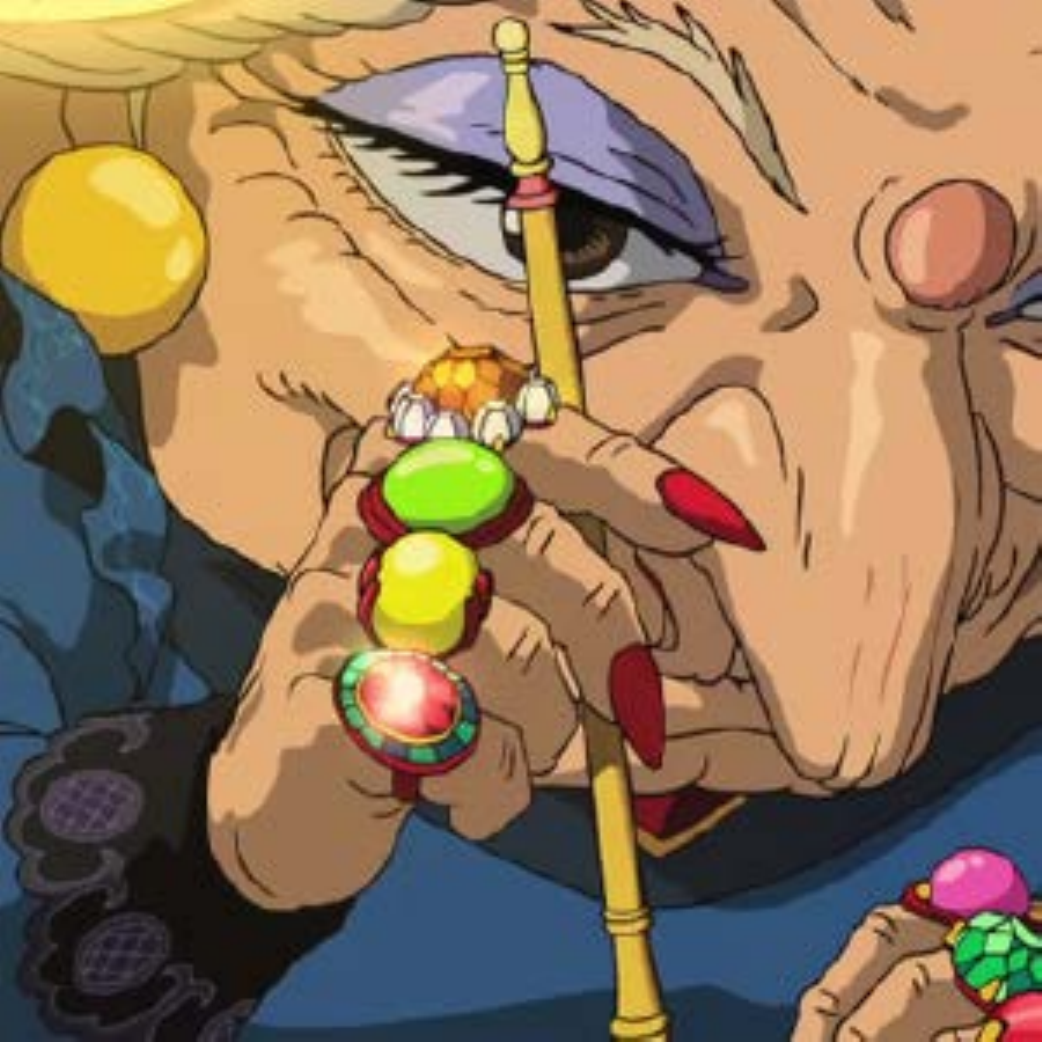} &
    \includegraphics[width=0.105\linewidth]{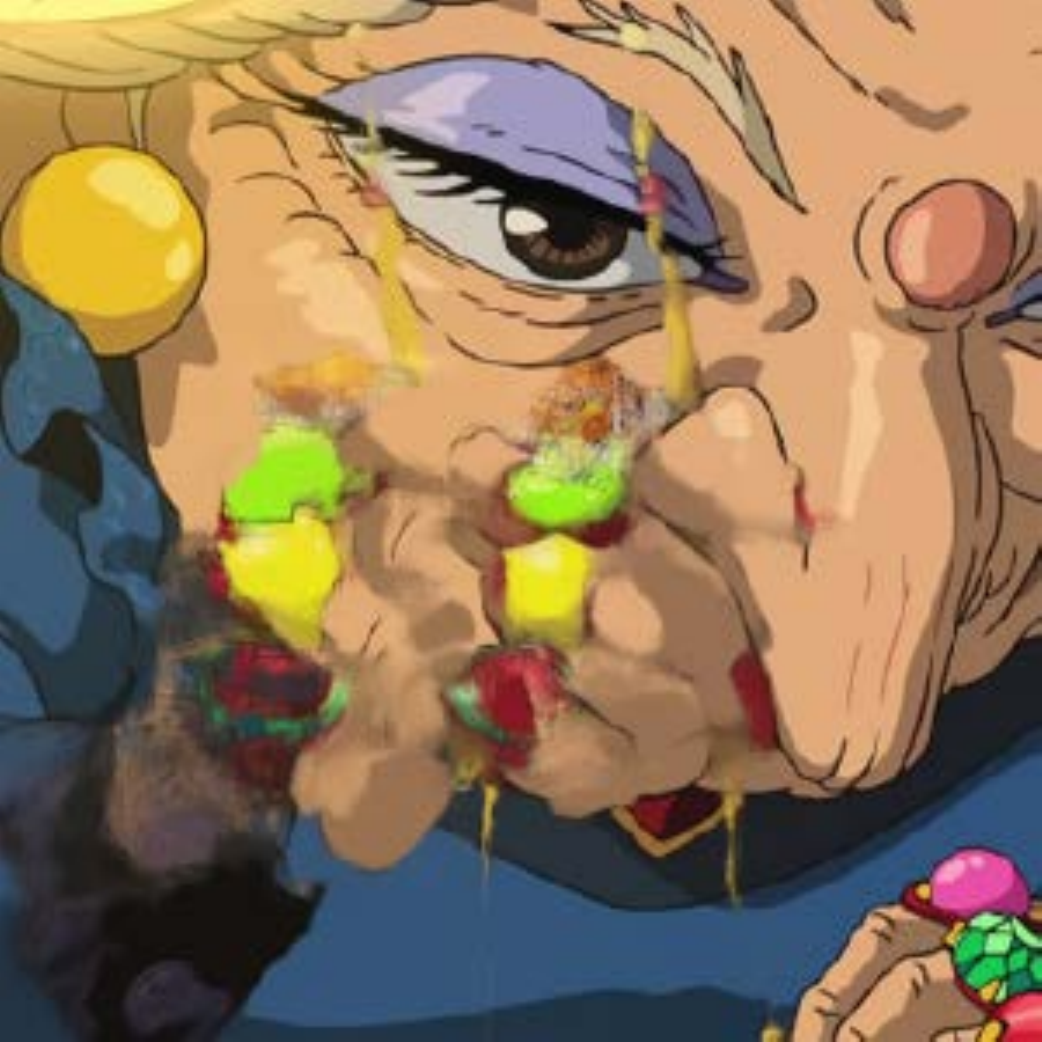}  &
    \includegraphics[width=0.105\linewidth]{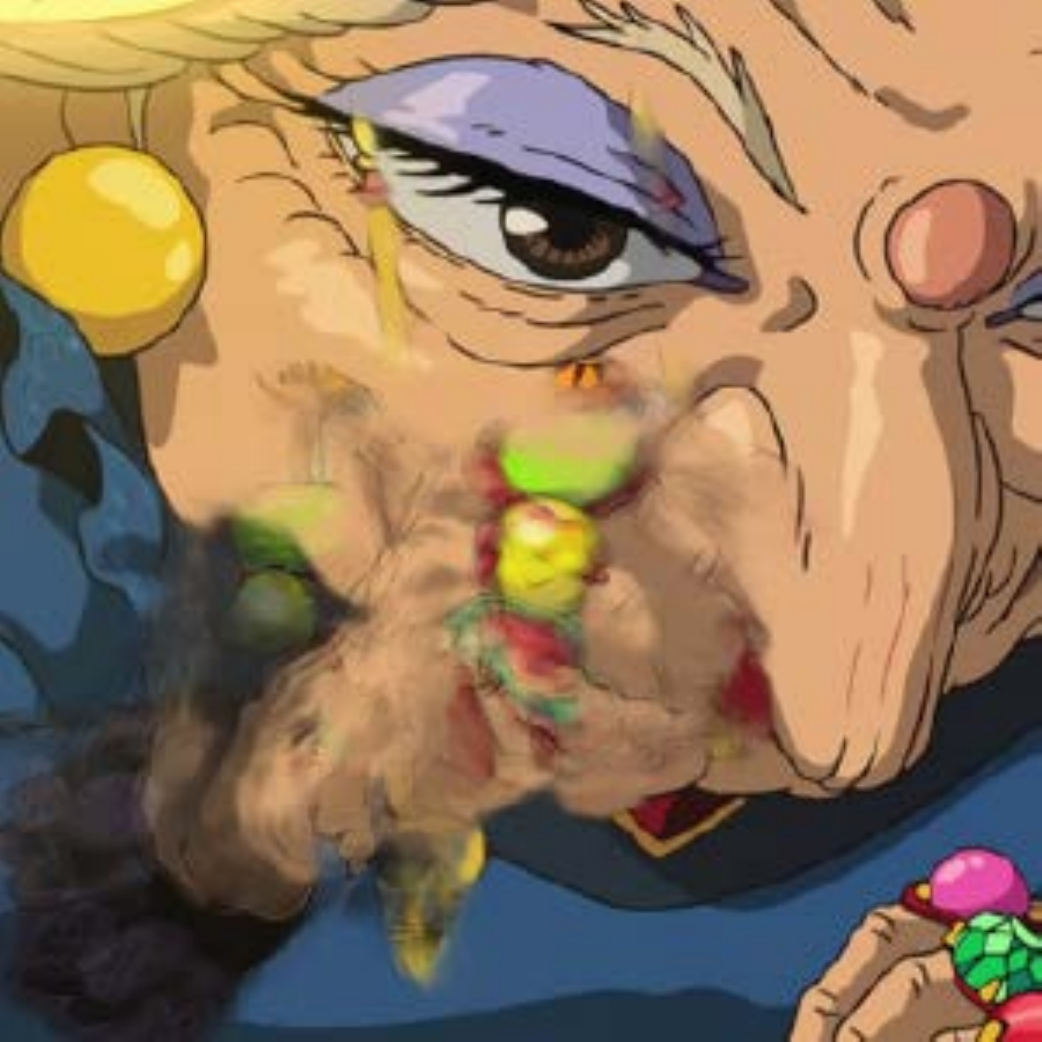}  &
    \includegraphics[width=0.105\linewidth]{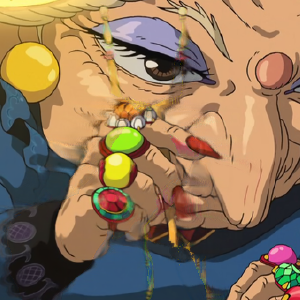}  &
    \includegraphics[width=0.105\linewidth]{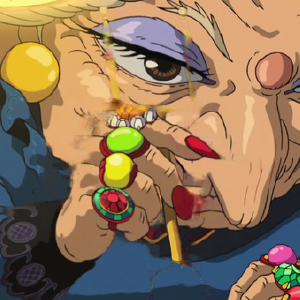}  &
    \includegraphics[width=0.105\linewidth]{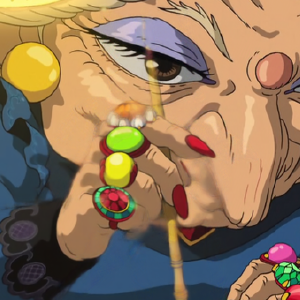}  &
    \includegraphics[width=0.105\linewidth]{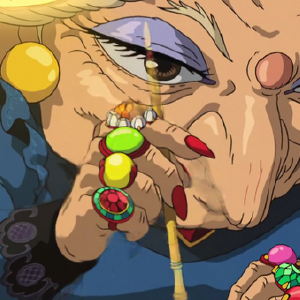}  \\

    Inputs &
    GT &
    BMBC~\cite{park2020bmbc} &
    AdaCoF~\cite{lee2020adacof} &
    DAIN~\cite{bao2019depth} &
    AutoFI\textit{ -dain} &
    ANIN~\cite{siyao2021deep} &
    AutoFI\textit{ -anin} \\
    
\end{tabular}
    \vspace{-5pt}
	\caption{
       Visual comparisons of AutoFI.
       The AutoFI synthesizes sharper and higher-quality interpolated frames.
	}
	\label{fig:compare_autofi_1}
	\vspace{-15pt}
\end{figure*}

\begin{table}[t]
	\vspace{-10pt}
\caption{Quantitative comparisons.}
\label{tab:compare}
\footnotesize
  \renewcommand{\tabcolsep}{10.0pt}
  \renewcommand{\arraystretch}{1.0} 
    \centering	
\begin{tabular}{lccc}
    \toprule
    Method               
    & PSNR $\uparrow$              
    & SSIM $\uparrow$
    & LPIPS $\downarrow$
    \\ 
    
    \midrule

    BMBC~\cite{park2020bmbc}
    & 27.77
    & 0.9146
    & 0.1025
    \\
    
    SepConv~\cite{niklaus2021revisiting}
    & 26.87
    & 0.9044
    & 0.1263
    \\
    
    AdaCoF~\cite{lee2020adacof}
    & 28.29
    & 0.9118
    & 0.1036
    \\

    DAIN~\cite{bao2019depth}          
    & 29.56
    & 0.9272
    & 0.0655
    \\
    
    ANIN~\cite{siyao2021deep}
    & 29.28
    & 0.9275 
    & 0.1036 
    \\
    
    \midrule

    AutoFI\textit{ -anin -$\mathcal{L}_{C}$ -w/o b }
    & 29.92
    & 0.9314
    & 0.0651
    \\

    AutoFI\textit{ -dain -$\mathcal{L}_{C}$}
    & 29.61 
    & 0.9278 
    & 0.0613
    \\

    AutoFI\textit{ -anin -$\mathcal{L}_{F}$}
    & {29.65} 
    & {0.9272} 
    & \underline{0.0611}
    \\

    AutoFI\textit{ -anin -$\mathcal{L}_{C}$}
    & \underline{30.10} 
    & \underline{0.9334} 
    & 0.0646
    \\

    \midrule

    SktFI\textit{ -none} 
    & 31.64
    & 0.9421
    & 0.0584
    \\

    SktFI\textit{ -dain} 
    &  31.90
    &  0.9470
    &  0.0506
    \\

    SktFI\textit{ -anin}
    &  \underline{32.22}
    &  \underline{0.9502}
    &  \underline{0.0486}
    \\

    \bottomrule
\end{tabular}
	\vspace{-10pt}
\end{table}
    
\subsection{Ablation Experiments}
    We conduct ablation experiments to evaluate the effectiveness of the proposed method.
    
    \Paragraph{Number of Training Steps.}
    As shown in~\figref{fig:psnr2}, running more iterations to train AutoFI\textit{ -anin}, such as $10^4$, results in moderate gains.
    When iterations come to $10^5$, the model is almost converged.

    \Paragraph{Loss Function.}
    The model with perceptual loss (AutoFI\textit{ -anin -$\mathcal{L}_{F}$}) achieve lower PSNR (0.45dB), but better LPIPS index than the model without using $\mathcal{L}_{F}$ (AutoFI\textit{ -anin -$\mathcal{L}_{C}$}), as shown in~\tabref{tab:compare}.
    The visual results in~\figref{fig:abla_percep_loss} show that the perceptual loss can help the model retain more details in difficult areas, resulting in sharper and higher quality frames. 
    
    \Paragraph{Gaussian Blur on AutoFI.}
    As shown in~\tabref{tab:compare}, Gaussian blur on AutoFI brings about $0.18$dB PSNR gain (AutoFI\textit{ -anin -$\mathcal{L}_{C}$} versus AutoFI\textit{ -anin -$\mathcal{L}_{C}$ -w/o b }).
    
    \Paragraph{Initial Frame on SktFI.}
    We compare the SktFI module without using initial interpolated frame (denoted by SktFI\textit{ -none}), using initial interpolated frame from AutoFI\textit{ -dain} (denoted by SktFI\textit{ -dain}) and using interpolated frame from AutoFI\textit{ -anin} (denoted by SktFI\textit{ -anin}).
    As shown in~\tabref{tab:compare}, SktFI\textit{ -anin} performs favorably against other methods, which means a better initial interpolated frame improve subsequent refinement performance.
    
    \begin{figure}[t]
	\footnotesize
	\centering
	\renewcommand{\tabcolsep}{1.0pt} 
	\renewcommand{\arraystretch}{0.5} 
	\begin{tabular}{ccc cc}

    \includegraphics[width=0.18\linewidth]{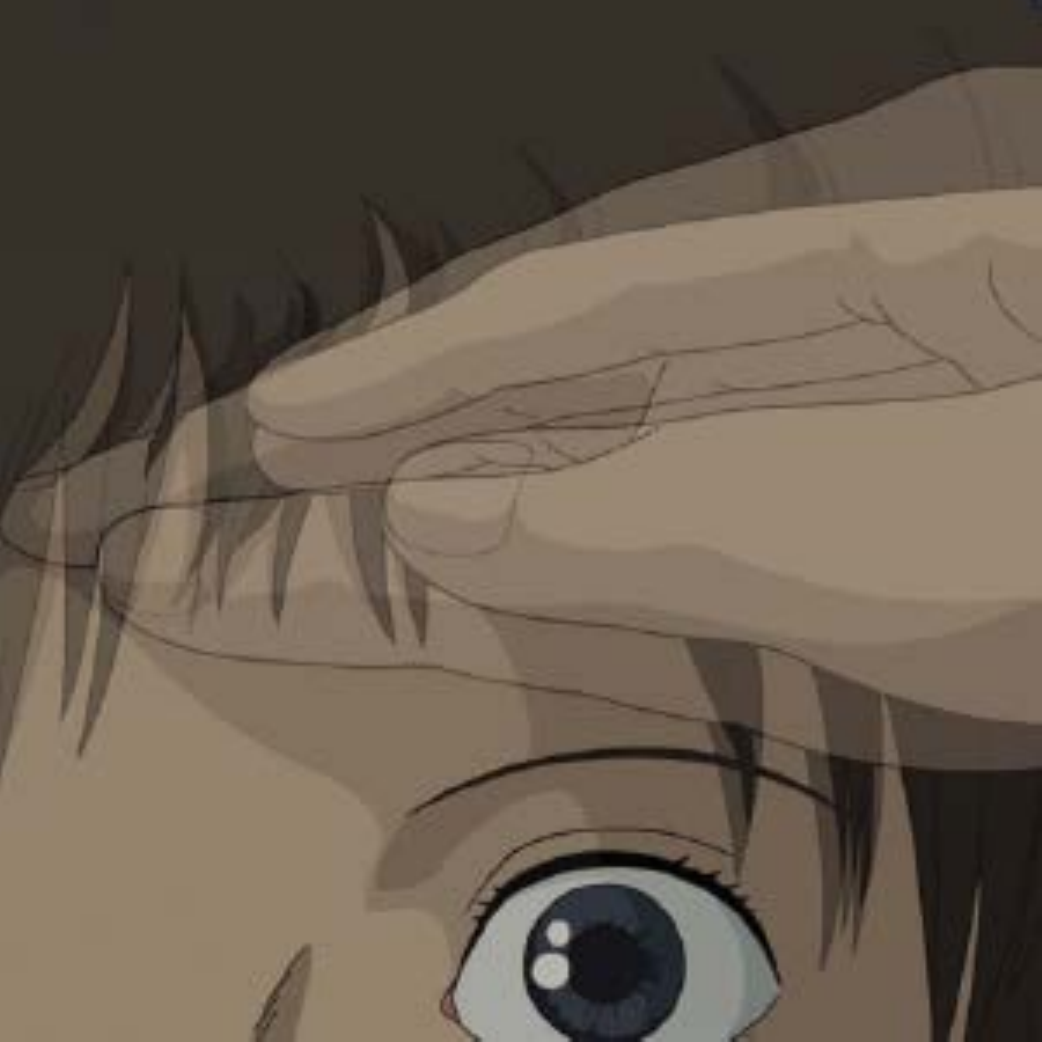}  &
    \includegraphics[width=0.18\linewidth]{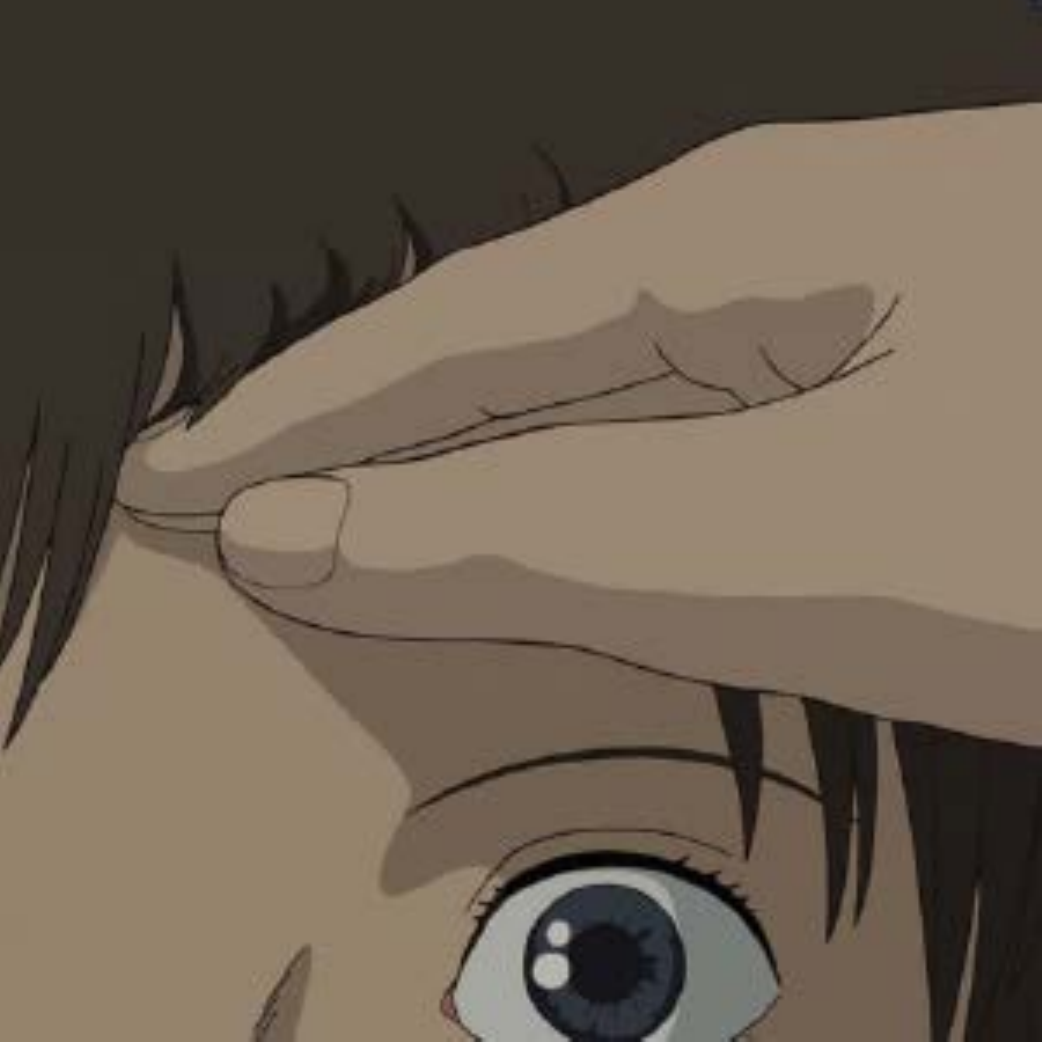} &
    \includegraphics[width=0.18\linewidth]{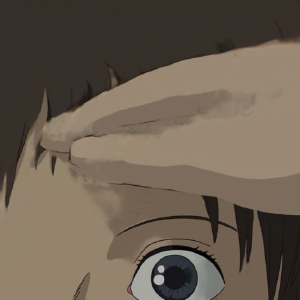} &
    \includegraphics[width=0.18\linewidth]{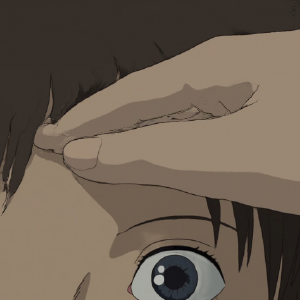} &
    \includegraphics[width=0.18\linewidth]{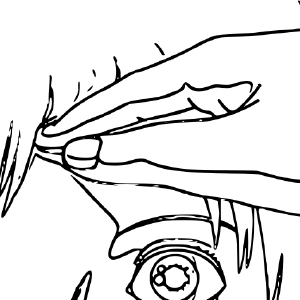}  \\

    \includegraphics[width=0.18\linewidth]{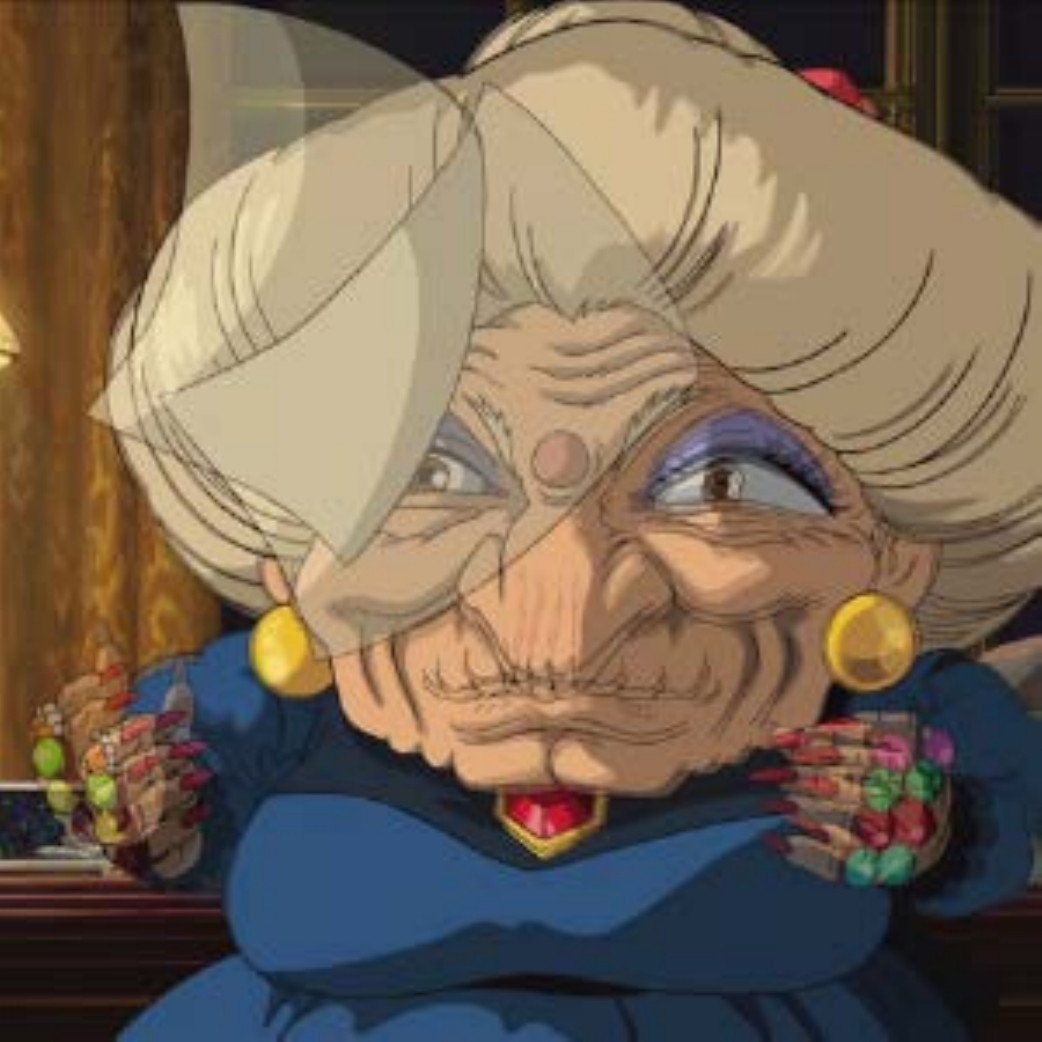}  &
    \includegraphics[width=0.18\linewidth]{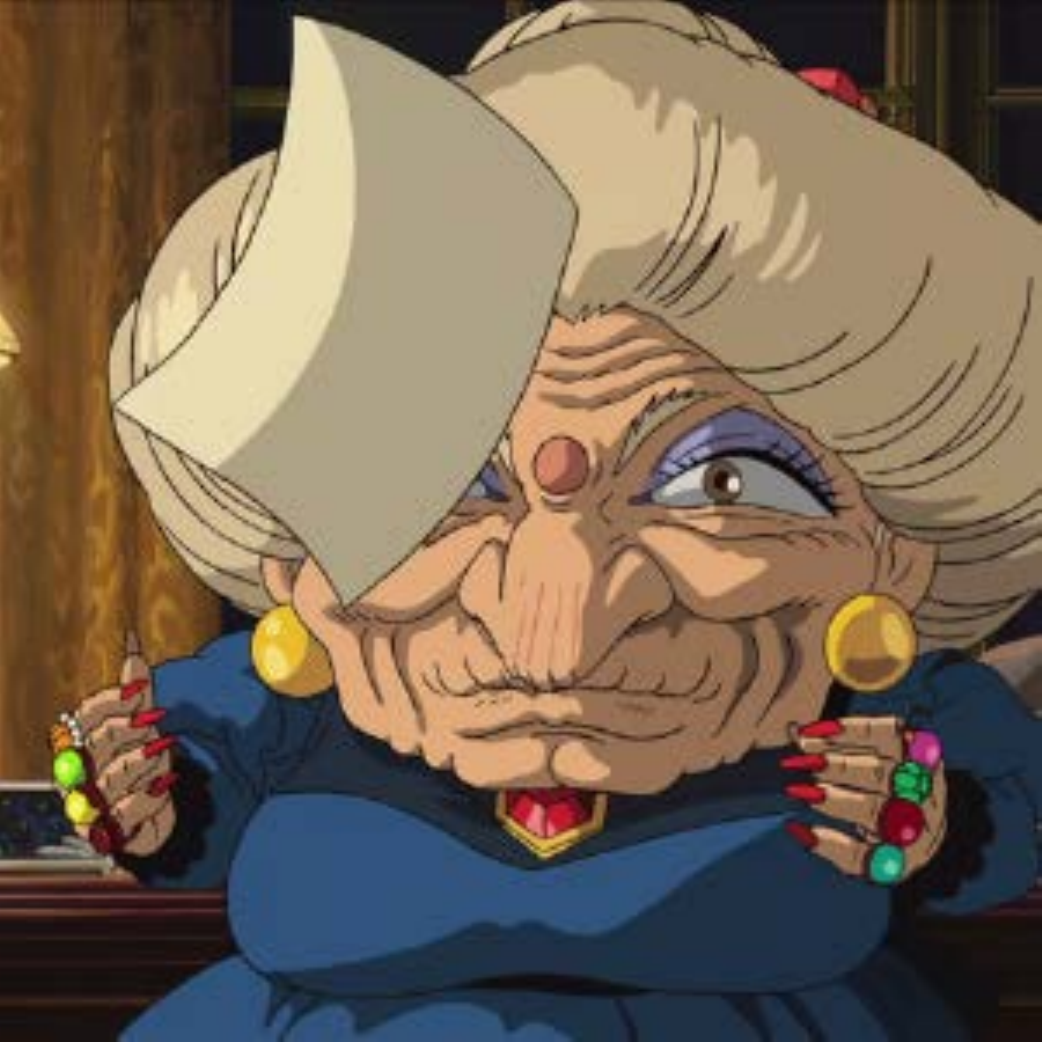} &
    \includegraphics[width=0.18\linewidth]{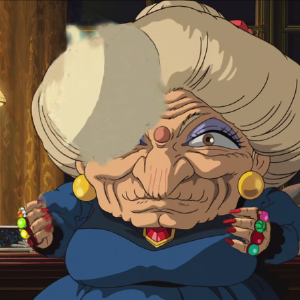} &
    \includegraphics[width=0.18\linewidth]{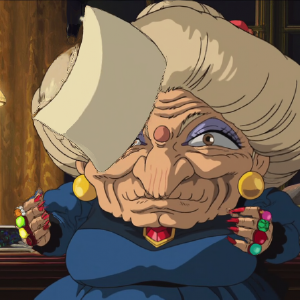} &
    \includegraphics[width=0.18\linewidth]{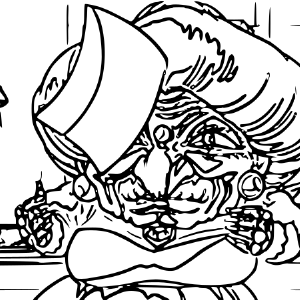}  \\

    \includegraphics[width=0.18\linewidth]{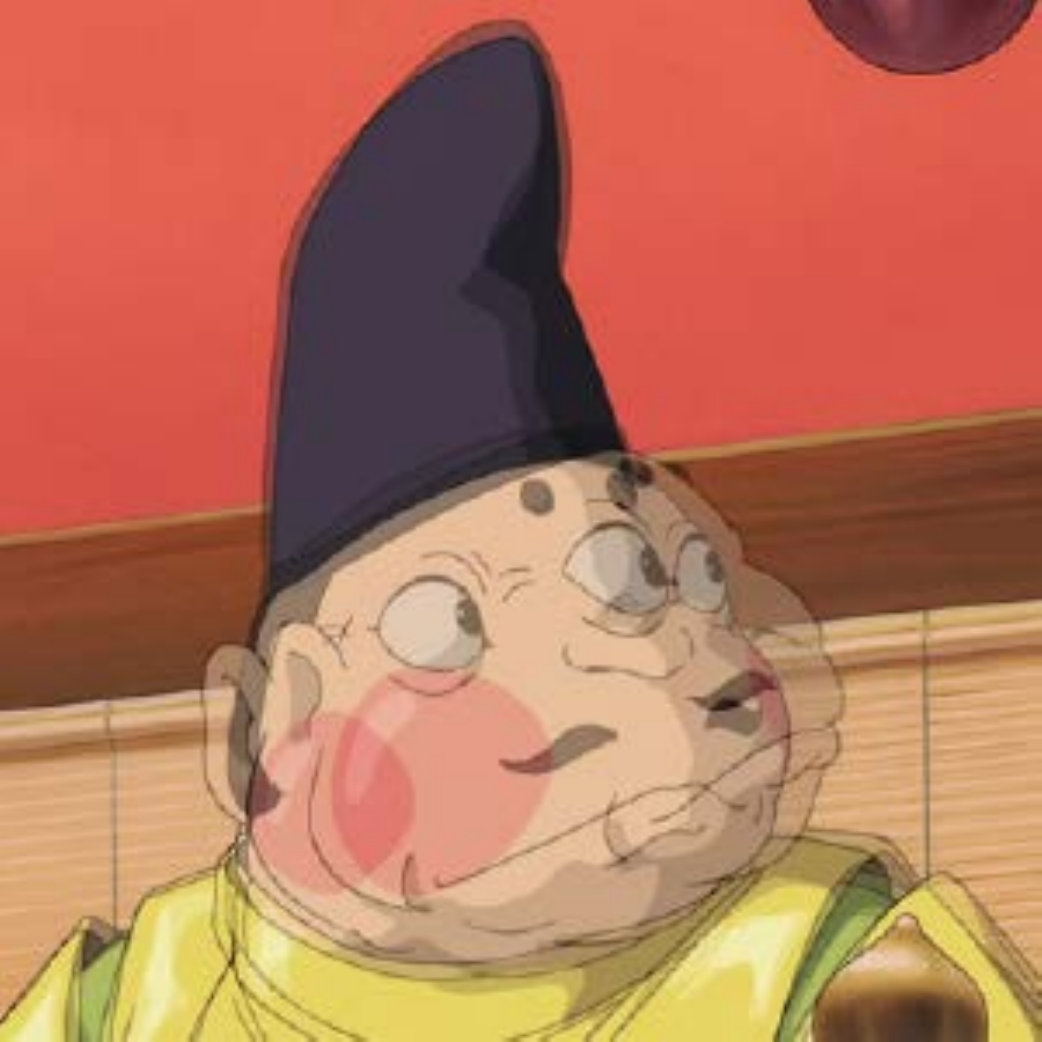}  &
    \includegraphics[width=0.18\linewidth]{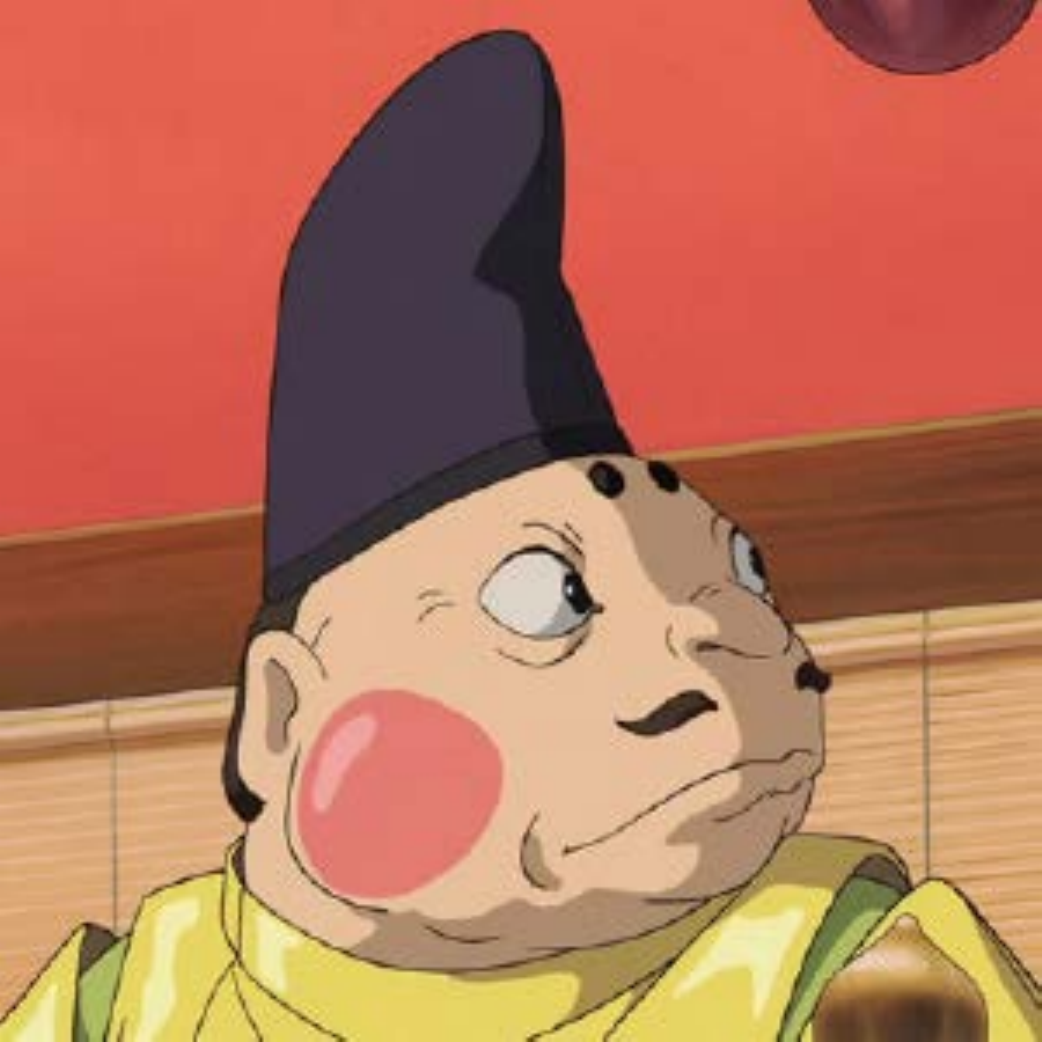} &
    \includegraphics[width=0.18\linewidth]{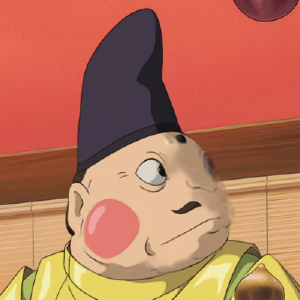} &
    \includegraphics[width=0.18\linewidth]{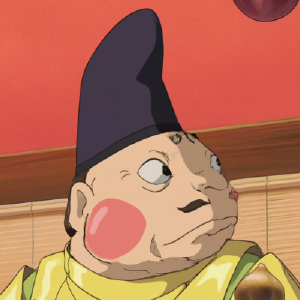} &
    \includegraphics[width=0.18\linewidth]{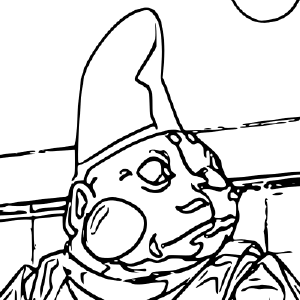}  \\

    Inputs &
    GT &
    AutoFI\textit{ -anin} & 
    SktFI\textit{ -anin} &
    Sketch \\
    
\end{tabular}
    \vspace{-7pt}
	\caption{
       Visual comparisons of SktFI.
      SktFI uses the sketch to refine the shape of the intermediate interpolated frame and restore more details.
	}
	\label{fig:compare_sktfi_visual}
	\vspace{-15pt}
\end{figure}

\subsection{Comparisons with State-of-the-arts}
    We compare the performance of AutoFI and SktFI with state-of-the-art frame interpolation baselines, including BMBC~\cite{park2020bmbc}, SepConv~\cite{niklaus2021revisiting}, AdaCoF~\cite{lee2020adacof}, DAIN~\cite{bao2019depth}, ANIN~\cite{siyao2021deep}.

    \Paragraph{Numerical Comparisons.}
    As shown in~\tabref{tab:compare}, the AutoFI\textit{ -anin -$\mathcal{L}_C$} outperforms all compared baselines, which obtains 0.82dB PSNR against ANIN~\cite{siyao2021deep}.
    Besides, the AutoFI\textit{ -anin -$\mathcal{L}_C$} achieves 0.49dB PSNR gain against AutoFI\textit{ -dain -$\mathcal{L}_C$}.
    The numerical results demonstrate the effectiveness of AutoFI.
    For the aspect of SktFI, as shown in~\tabref{tab:compare}, the Skt\textit{ -anin} has best interpolation performance, which obtains about 2dB PSNR gain against  AutoFI\textit{ -anin -$\mathcal{L}_C$}.
    The main reason for the high performance of SktFI is that the sketch is drawn from the ground-truth intermediate frame of the test set with nonlinear motion.
    In contrast, frame interpolation algorithms are under the linear motion assumption. 
    Thus the sketch rectifies the linear interpolated frame to a nonlinear version, which fits the non-linearity of the test set.

    \Paragraph{Visual Comparisons.}
    We show visual comparisons of AutoFI in~\figref{fig:compare_autofi_1}.
    The AutoFI synthesizes sharper and higher-quality interpolated frames.
    AutoFI ensures the basic assumption of linear motion of frame interpolation algorithms.
    While the non-linearity of the traditional training set, such as ATD-12K~\cite{siyao2021deep} misleads the frame interpolation network, resulting in blurry results.
    We show visual comparisons of SktFI in~\figref{fig:compare_sktfi_visual}.
    The user-provided sketch in SktFI provides the cues of intermediate contours, helping to refine the shape of the intermediate interpolated frame and restore more details.

\section{Conclusion}

    This paper presents AutoFI, an effective method to render training data for deep animation video interpolation.
    AutoFI takes a layered architecture to render synthetic data, which ensures the assumption of linear motion.
    Besides AutoFI, we also propose a plug-and-play sketch-based post-processing module, named SktFI, to refine the final results using user-provided sketches manually.
    AutoFI and SktFI help to improve the animation production pipeline.
    Numerical and visual results demonstrate that the interpolated animation frames show high-performance and high-perceptual-quality.


\clearpage

\bibliographystyle{IEEEbib}
\bibliography{strings,refs}

\end{document}